\documentclass[journal]{IEEEtran}

\IEEEoverridecommandlockouts
\usepackage{cite}
\usepackage{amsmath,amssymb,amsfonts}
\usepackage{algorithmic}
\usepackage{graphicx}
\usepackage{textcomp}
\usepackage{xcolor}
\usepackage{tikz}
\usepackage{standalone}
\def\BibTeX{{\rm B\kern-.05em{\sc i\kern-.025em b}\kern-.08em
    T\kern-.1667em\lower.7ex\hbox{E}\kern-.125emX}}

\newcommand{\algcomment}[1]{\textcolor{blue}{$\triangleleft\quad$\textit{#1}$\quad\triangleright$}}
\usepackage{ml_package}

\newcommand{\changemade}[1]{{#1}}

\begin{document}

\title{Bayesian~Active~Meta-Learning for Reliable and Efficient AI-Based~Demodulation}

\author{    
    Kfir~M.~Cohen,~\IEEEmembership{Student Member,~IEEE}, Sangwoo~Park,~\IEEEmembership{Member,~IEEE},\\           Osvaldo~Simeone,~\IEEEmembership{Fellow,~IEEE},~Shlomo~Shamai~(Shitz),~\IEEEmembership{Life~Fellow,~IEEE}
     % <-this % stops a space % don't delete the following blank line if you want names to be centered

\thanks{Part of this work was presented in WSA 2021 - 25th International ITG Workshop on Smart Antennas \cite{cohen2021learning}.}
\thanks{The work of K. M. Cohen, S. Park and O. Simeone has been supported by the European Research Council (ERC) under the European Union’s Horizon 2020 research and innovation programme, grant agreement No. 725731. The work of O. Simeone has also been supported by an Open Fellowship of the EPSRC.}
\thanks{The work of S. Shamai has been supported by the European Union's Horizon 2020 Research And Innovation Programme, grant agreement No. 694630.}
\thanks{The authors acknowledge use of the research computing facility at King's College London (2022): King's Computational Research, Engineering and Technology Environment (CREATE). Retrieved October 25, 2022, from https://doi.org/10.18742/rnvf-m076; and Rosalind (https://rosalind.kcl.ac.uk)}
\thanks{Kfir M. Cohen, Sangwoo Park, and Osvaldo Simeone are with King's Communication, Learning, \& Information Processing (KCLIP) lab, Department of Engineering, King’s College London, London WC2R 2LS, U.K. (e-mail: kfir.cohen@kcl.ac.uk; sangwoo.park@kcl.ac.uk; osvaldo.simeone@kcl.ac.uk).}
\thanks{Shlomo Shamai (Shitz) is with the Viterbi Faculty of Electrical and Computing Engineering, Technion—Israel Institute of Technology, Haifa, Israel
3200003 (e-mail: sshlomo@ee.technion.ac.il).}
\thanks{Code for this work can be found in \texttt{https://github.com/kclip/bayesian\_active\_meta\_learning}.}
}

\maketitle

\begin{abstract}

Two of the main principles underlying the life cycle of an artificial intelligence (AI) module in communication networks are \emph{adaptation} and \emph{monitoring}. Adaptation refers to the need to adjust the operation of an AI module depending on the current conditions; while monitoring requires measures of the reliability of an AI module's decisions. Classical frequentist learning methods for the design of AI modules fall short on both counts of adaptation and monitoring, catering to one-off training and  providing overconfident decisions. This paper proposes a solution to address both challenges by integrating meta-learning with Bayesian learning. As a specific use case, the problems of demodulation and equalization over a fading channel based on the availability of few pilots are studied. Meta-learning processes pilot information from multiple frames in order to extract useful shared properties of effective demodulators across frames. The resulting trained demodulators  are demonstrated, via experiments, to offer better calibrated soft decisions, at the computational cost of running an ensemble of networks at run time. The capacity to quantify uncertainty in the model parameter space is further leveraged by extending Bayesian meta-learning to an active setting. In it, the designer can select in a sequential fashion channel conditions under which to generate data for meta-learning from a channel simulator. Bayesian active meta-learning is seen in experiments to significantly reduce the number of frames required to obtain efficient adaptation procedure for new frames.
\end{abstract}

\begin{IEEEkeywords}
Bayesian meta-learning, uncertainty quantification, Bayesian active meta-learning, demodulation.
\end{IEEEkeywords}

%%%%%%%%%%%%%%%%%%%%%%%%%%%%%%%%%%%%%%%%%%%%%%%%%%%%%%%%%%%%%%%%%%%%%%%%%%%%%%%%%%%%%%%%%%%%%%%%%%%%%%%%%%
%%%%%%%%%%%%%%%%%%%%%%%%%%%%%%%%%%%%%%          Section             %%%%%%%%%%%%%%%%%%%%%%%%%%%%%%%%%%%%%%
%%%%%%%%%%%%%%%%%%%%%%%%%%%%%%%%%%%%%%%%%%%%%%%%%%%%%%%%%%%%%%%%%%%%%%%%%%%%%%%%%%%%%%%%%%%%%%%%%%%%%%%%%%
\section{Introduction}
%%%%%%%%%%%%%%%%%%%%%%%%%%%%%%%%%%%%%%%%%%%%%%%%%%%%%%%%%%%%%%%%%%%%%%%%%%%%%%%%%%%%%%
%%%%%%%%%%%%%%%%%%%%%%%%%%%%        Subsection            %%%%%%%%%%%%%%%%%%%%%%%%%%%%
%%%%%%%%%%%%%%%%%%%%%%%%%%%%%%%%%%%%%%%%%%%%%%%%%%%%%%%%%%%%%%%%%%%%%%%%%%%%%%%%%%%%%%
\subsection{ Motivation}

Artificial intelligence (AI) is seen as a key enabler for next-generation wireless systems \cite{letaief2019roadmap}. Emerging solutions, such as Open-Radio Access Network (O-RAN), incorporate AI modules as native components of a modular architecture that can be fine-tuned to meet the requirements of specific deployments \cite{bonati2021intelligence}. Two of the main principles underlying the life cycle of an AI module in communication networks are \emph{adaptation} and \emph{monitoring} \cite{masur2022artificial}. Adaptation refers to the need to adjust the operation of an AI module depending on the current conditions, particularly for real-time applications at the frame level. At run time, an AI model should ideally enable monitoring of the quality of its outputs by providing measures of the reliability of its decisions. The availability of such reliability measures is instrumental in supporting several important functionalities, from the combination of multiple models to decisions about retraining \cite{oran2020}. 

Classical frequentist learning methods for the design of AI modules fall short on both counts of adaptation and monitoring (see, e.g., \cite{simeone2020learning,Guo2017Calibration}). First, conventional frequentist learning is well known to provide inaccurate measures of reliability, typically producing overconfident decisions \cite{Guo2017Calibration}. Second, the standard learning approach prescribes the one-off optimization of an AI model, hence failing to capture the need for adaptation. This paper investigates the integration of meta-learning and Bayesian learning as a means to address both challenges. As we detail in the next section, Bayesian learning can provide well-calibrated, and hence reliable, measures of uncertainty of a model’s decision; while meta-learning can reduce the amount of data required for adaptation to a new task, thus improving efficiency. As a specific use case, we focus on the problems of demodulation and equalization over a fading channel based on the availability of few pilots (see Fig.~\ref{fig: multi frames to base station demodulation problem}). The goal is to develop AI solutions that are capable of adapting the demodulator/equalizer to changing conditions based on few training symbols, while also being able to quantify the uncertainty of the AI model's output.

%%%%%%%%%%%%%%%%%%%%%%%%%%%%%%%%%%%%%%%%%%%%%%%%%%%%%%%%%%%%%%%%%%%%%%%%%%%%%%%%%%%%%%
%%%%%%%%%%%%%%%%%%%%%%%%%%%%        Subsection            %%%%%%%%%%%%%%%%%%%%%%%%%%%%
%%%%%%%%%%%%%%%%%%%%%%%%%%%%%%%%%%%%%%%%%%%%%%%%%%%%%%%%%%%%%%%%%%%%%%%%%%%%%%%%%%%%%%
\subsection{Background}

As illustrated in Fig. \ref{Fig: Evolution of network, determinitstic to Bayesian amortized VI}, frequentist learning assigns a single value to each model parameter as a result of training. This neglects (epistemic) uncertainty that exists at the level of model parameters due to the limited availability of data. In contrast, \textit{Bayesian learning} can express uncertainty about the true value of the model parameter vector by optimizing over a distribution, rather than over a single point value \cite{Barber2012book}. By averaging predictions over the distribution of the model parameters, Bayesian learning is known to be capable of providing decisions that are well calibrated \cite{blundell2015weight, wang2020survey, gal2016dropout}. Calibration refers to the capacity of a model to produce confidence levels that reproduce well the actual accuracy of the decisions.

\begin{figure}
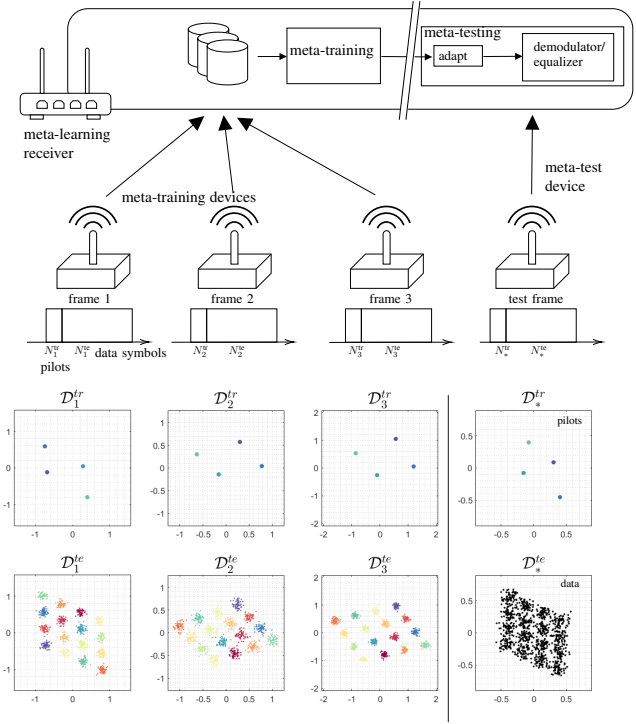

    \centering
    \includestandalone[trim = 1.5cm 0cm 0cm 0cm,clip,width=9cm]{fig_tikz_datasets_mtr_mte_qam}%     without .tex extension  % this one uses fig_datasets_meta_training_and_testing.png inside the tex code
    \caption{Illustration of the meta-learning problem studied in this work for the example of 16-ary quadrature amplitude modulation (16-QAM). A receiver has available data corresponding to frames previously received from multiple devices, each possibly experiencing different channel conditions. Given meta-training data sets $\{\dataset_\tau\}_{\tau=1}^t$ of pilots from previous frames, partitioned into training data and test data, the demodulator optimizes a hyperparameter vector $\xi$. For a newly received frame, the receiver uses the few pilots therein to adapt the demodulator/equalizer parameter vector $\phi_*$. In the Bayesian meta-learning framework, instead of a single parameter vector $\phi_*$, the receiver optimizes over an ensemble of parameter vectors through the hyperparameter vector $\xi$ of a posterior distribution $p(\phi_*|\dataset_*^\text{tr},\xi)$.}
    \label{fig: multi frames to base station demodulation problem}
\end{figure}

\emph{Meta-learning}, also known as \emph{learning to learn}, optimizes training strategies that can fine-tune a model based on few samples for a new task by transferring knowledge across different learning tasks \cite{thrun1998lifelong,Finn2017maml, zintgraf2019fast, maclaurin2015gradient, li2017meta, behl2019alpha,baxter1998theoretical}. Meta-learning is a natural tool to produce AI solutions that are optimized for adaptation. Prior work on meta-learning for communication systems, including \cite{park2021fewpilots, goutay2020deep, yuan2020transfer, hu2020meta, raviv2021meta, kalor2021latency, nikoloska2021fast, zhang2021embedding, kalor2021prediction, jiang2019mind, zhang2020meta}, is limited to standard frequentist learning. Therefore, existing art is unable to produce models that provide well-calibrated estimations of reliability. Most related to our work is \cite{park2021fewpilots}, which proposes to leverage pilot information from previous frames in order to optimize training procedures to be applied to the pilots of new frames (see Fig.~\ref{fig: multi frames to base station demodulation problem}).

\emph{Bayesian meta-learning} aims at optimizing the procedure that produces the posterior distribution for new learning tasks. Accordingly, the goal of Bayesian meta-learning is to enhance the efficiency of Bayesian learning by reducing the number of training points needed to obtain accurate and well calibrated Bayesian models. The optimization of the Bayesian learning process is carried out by transferring knowledge from previously encountered tasks for which data are assumed to be available \cite{nguyen2020uncertainty,posch2019variational,Sun2021amortized}. To the best of our knowledge, with the exception of the conference version of this paper \cite{cohen2021learning}, this is the first work to consider the application of Bayesian meta-learning to communication systems.

Beside meta-learning, another approach to reduce the number of required training data points is \emph{active learning} \cite{houlsby2011bayesian,gal2017deep,sohrabi2022active,sahbi2021active,kaddour2020probabilistic}. Active learning amounts to the process of choosing which samples should be annotated next and incrementally added to the training set \cite{kirsch2019batchbald}. Through this process, active learning can select relevant samples at which the model is currently most uncertain in order to speed up the training process.

A much less studied area is \emph{active meta-learning}, which aims at reducing the number of tasks a meta-learner must collect data from, before it can adapt efficiently to new tasks \cite{kaddour2020probabilistic,nikoloska2021bamldbo}. Reference \cite{kaddour2020probabilistic} proposes a method based on Bayesian meta-learning via empirical Bayes; while the paper \cite{nikoloska2021bamldbo} takes a hierarchical Bayesian approach, generalizing the Bayesian active learning by disagreements (BALD) criterion introduced in \cite{houlsby2011bayesian} to meta-learning. While \cite{kaddour2020probabilistic} assumes labeled training sets, reference \cite{nikoloska2021bamldbo} considers unlabeled data during active meta-learning. As such, the setting in it is not applicable to the problem under study here in which data consists of supervised pairs of pilots and received signals (see Fig.~\ref{fig: multi frames to base station demodulation problem}). A summary of the relevant approaches built upon in this work is given in Table~\ref{tab: Built Upon Approaches Enabling Reliability and Sample-Efficiency}.

\begin{table*}[tbp]
    \centering
    \caption{A Summary of the relevant techniques considered in this work}
    \label{tab: Built Upon Approaches Enabling Reliability and Sample-Efficiency}
    \begin{tabular}{ p{0.16\linewidth} p{0.22\linewidth} p{0.50\linewidth} } \toprule
        Approach
        & Goal
        & Methodology
        \\ [0.5ex] \midrule\midrule
        frequentist learning
        & accurate data-driven predictions
        & minimize the training loss over the model parameter vector $\phi$
        \\ [0.5ex] \hline
        \vtop{\hbox{\strut Bayesian learning}\hbox{\strut via variational inference}}
        & \vtop{\hbox{\strut reliable and accurate}\hbox{\strut data-driven predictions}}
        & minimize the free energy over the parameters $\varphi$ of the variational distribution $q(\phi|\varphi)$
        \\ [0.5ex] \hline
        frequentist meta-learning
        & sample efficiency
        & minimize the meta-training loss over hyperparameters $\xi$ to be used for frequentist learning
        \\ [0.5ex] \hline
        Bayesian meta-learning
        & sample efficiency and reliability
        & minimize the meta-training loss over hyperparameters $\xi$ to be used for Bayesian learning
        \\ [0.5ex] \hline
        \vtop{\hbox{\strut Bayesian active}\hbox{\strut meta-learning}}
        & task efficiency, sample efficiency, and reliability
        & minimize the meta-training loss over the hyperparameters $\xi$ and over the sequential selection of meta-learning tasks
        \\% [0.5ex] 
        \bottomrule
    \end{tabular}
    \vspace{0.2cm}
\end{table*}

\begin{figure*}%[!ht]
    \centering
    \includegraphics[page=1,trim = 0cm 0.3cm 0cm 0cm, clip, width=0.85\textwidth]{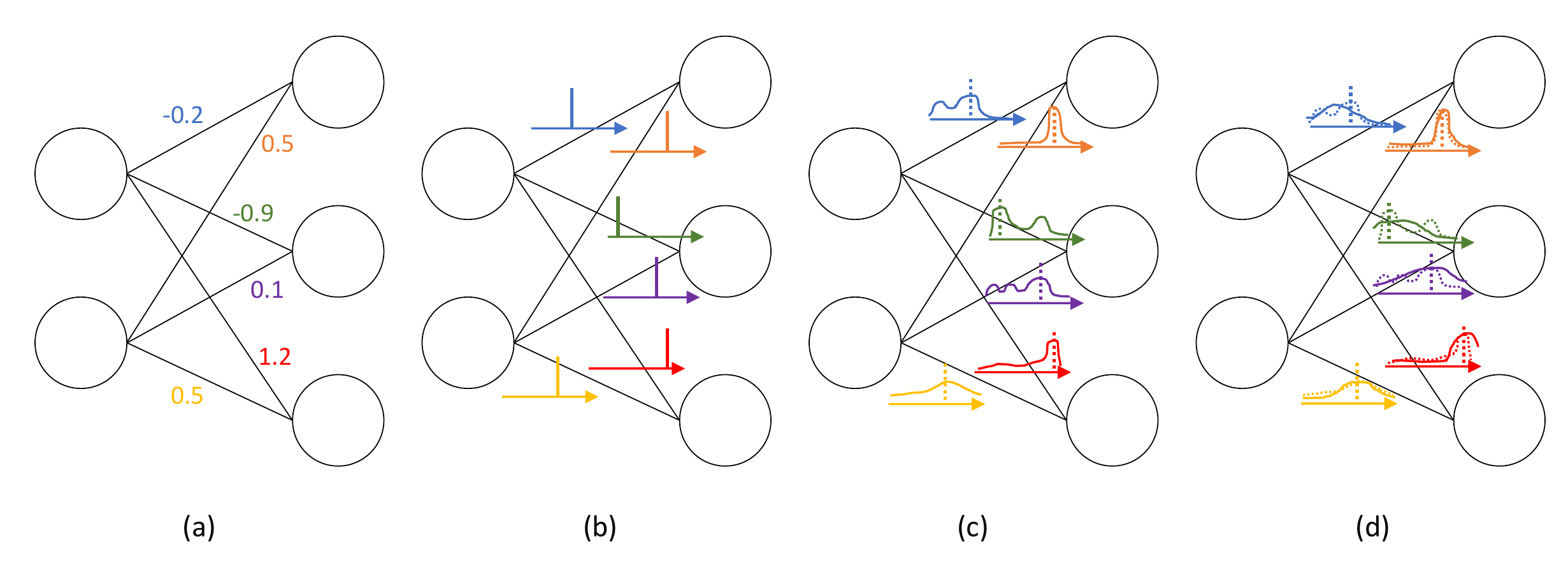}
    \caption{Network weights in frequentist and Bayesian learning: (a) in frequentist learning, each weight is described by a scalar value; (b) the scalar value can be viewed as random variable having a degenerated probabilistic distribution concentrated at a simple prior; (c) in Bayesian learning, the weights are assigned a probability distribution, which,  unlike the frequentist point estimate (dashed vertical line), provides information about the uncertainty on the weight; (d) in variational inference (VI), the posterior is approximated with a parameter distribution.}
    \label{Fig: Evolution of network, determinitstic to Bayesian amortized VI}
\end{figure*}

%%%%%%%%%%%%%%%%%%%%%%%%%%%%%%%%%%%%%%%%%%%%%%%%%%%%%%%%%%%%%%%%%%%%%%%%%%%%%%%%%%%%%%
%%%%%%%%%%%%%%%%%%%%%%%%%%%%        Subsection            %%%%%%%%%%%%%%%%%%%%%%%%%%%%
%%%%%%%%%%%%%%%%%%%%%%%%%%%%%%%%%%%%%%%%%%%%%%%%%%%%%%%%%%%%%%%%%%%%%%%%%%%%%%%%%%%%%%
\subsection{Contributions}
This paper introduces the use of Bayesian meta-learning to enable both adaptation and monitoring for the tasks of demodulation and equalization. Unlike prior works that considered either frequentist meta-learning \cite{park2021fewpilots, simeone2020learning, goutay2020deep, yuan2020transfer, hu2020meta, raviv2021meta, kalor2021latency, nikoloska2021fast, zhang2021embedding} or Bayesian learning \cite{zhang2017bayesian,zhang2017novel,prasad2015joint,lv2019joint}, the proposed Bayesian meta-learning methodology enables both resource-efficient adaptation and a reliable quantification of uncertainty. To further improve the efficiency of Bayesian meta-learning we propose the use of active meta-learning, which reduces the number of required meta-training data from previously received frames. Specific contributions are as follows.

\begin{itemize}
    \item We introduce Bayesian meta-learning for the problems of demodulation and equalization from few pilots. The proposed implementation is derived based on parametric VI.
    \item We introduce Bayesian active meta-learning as a solution to reduce the number of frames required by meta-learning. Active meta-learning selects in a sequential fashion channel conditions under which to generate data for meta-learning from a channel simulator.
    \item Extensive experimental results demonstrate that Bayesian meta-learning produces demodulators and equalizers that offer better calibrated soft decisions. Furthermore, they show that for a target meta-testing loss, active meta-learning can reduce the number of simulated meta-training frames required.
\end{itemize}
Part of this paper was presented in \cite{cohen2021learning}, which presented the idea of Bayesian meta-learning with some preliminary experiments. This journal version presents full technical details, new results and introduces for the first time Bayesian active meta-learning for communication systems.

The rest of the paper is organized as follows. Section \ref{sec: channel model} introduces the channel model, along with background material on standard frequentist learning and frequentist meta-learning. Section \ref{sec: Bayesian VI ML} expands on Bayesian meta-learning. Then, we present Bayesian active meta-learning in Section \ref{sec: Bayesian Active Meta-Learning}. Numerical results are presented in Section \ref{sec: Experiments}, and Section \ref{sec: conclusions} concludes the paper.

%%%%%%%%%%%%%%%%%%%%%%%%%%%%%%%%%%%%%%%%%%%%%%%%%%%%%%%%%%%%%%%%%%%%%%%%%%%%%%%%%%%%%%
%%%%%%%%%%%%%%%%%%%%%%%%%%%%        Subsection            %%%%%%%%%%%%%%%%%%%%%%%%%%%%
%%%%%%%%%%%%%%%%%%%%%%%%%%%%%%%%%%%%%%%%%%%%%%%%%%%%%%%%%%%%%%%%%%%%%%%%%%%%%%%%%%%%%%
\subsection{Related Work}

For scalability, Bayesian learning can be implemented via approximate methods based on variational inference (VI) or Monte Carlo (MC). VI methods approximate the exact Bayesian posterior distribution with a tractable variational density \cite{blundell2015weight,graves2011practical,dusenberry2020efficient,daxberger2021laplace,farquhar2020liberty}, while Monte Carlo techniques obtain approximate samples from the Bayesian posterior distribution \cite{neal2011mcmc,welling2011bayesian,zhang2019cyclical}. Each class of methods comes with its own set of technical challenges and engineering choices. For instance, VI requires the selection of a variational distribution family, such as mean-field Gaussian models, and the specification of a stochastic optimization algorithm. There are also non-parametric VI methods such as Stein variational gradient descent (SVGD) \cite{liu2016stein}, which optimize over deterministic and interacting particles. For MC techniques, solutions range from first-order Langevin dynamics techniques \cite{welling2011bayesian} to more complex methods such as Hamiltonian Monte Carlo (HMC) \cite{neal2011mcmc}. Implementing any of these schemes for a specific engineering application is a non-trivial task.

Bayesian learning has been applied in reference \cite{narmanlioglu2017prediction} to the problem of predicting the number of active users in LTE system; papers \cite{wu2022stochastic,zilberstein2022annealed} applied MC-based Bayesian learning for MIMO detection; the works \cite{tao2021improved,jha2021transformer,jha2021online} addressed channel prediction/estimation for massive MIMO systems; reference \cite{xu2021bayesian} studied the identification of IoT transmitters; and the authors of \cite{zecchin2022robust} proposed the use of robust Bayesian learning for modulation classification, localization, and channel modeling. %Bayesian learning has been applied to communication systems in \cite{zhang2017bayesian,zhang2017novel,prasad2015joint,lv2019joint,Maggi2021}.

As for active learning, applications to communication systems include paper \cite{liu2020wireless}, which proposed a sample-efficient retransmission protocol; reference \cite{chiu2019active}, which tackled initial beam alignment for massive MIMO system; work \cite{yang2018active}, which aimed at mitigating the problem of scarce training data in wireless cyber-security attack; and reference \cite{abdel2019ultra}, which addressed resource allocation problems in vehicular communication systems.

Like Bayesian learning, meta-learning also provides a general design principle, which can be implemented by following different approaches. Optimization-based methods design the hyperparameters used by training algorithms; model-based techniques optimize an additional neural network model to guide adaptation of the main AI model; and metric-based schemes identify metric spaces for non-parametric inference (see, e.g., \cite{hospedales2021meta} and references therein).

The integration of meta-learning and Bayesian learning is highly non-trivial, and is an active topic of research in the machine learning literature.
References \cite{finn2018probabilistic, ravi2018amortized, yoon2018bayesian,patacchiola2020bayesian, zou2020gradient} addressed Bayesian meta-learning via empirical Bayes using parametric VI \cite{finn2018probabilistic, ravi2018amortized}, particle-based VI \cite{yoon2018bayesian}, deep-kernels \cite{patacchiola2020bayesian}, and expectation-maximization \cite{zou2020gradient}; while the papers \cite{amit2018meta, rothfuss2021pacoh, rothfuss2021meta, jose2022information} studied full Bayesian meta-learning that treats also the hyperparameters as random variables. Lastly, the work \cite{nikoloska2022quantum} proposed the use of quantum machine learning models as parameterized variational distributions.

%%%%%%%%%%%%%%%%%%%%%%%%%%%%%%%%%%%%%%%%%%%%%%%%%%%%%%%%%%%%%%%%%%%%%%%%%%%%%%%%%%%%%%%%%%%%%%%%%%%%%%%%%%
%%%%%%%%%%%%%%%%%%%%%%%%%%%%%%%%%%%%%%          Section             %%%%%%%%%%%%%%%%%%%%%%%%%%%%%%%%%%%%%%
%%%%%%%%%%%%%%%%%%%%%%%%%%%%%%%%%%%%%%%%%%%%%%%%%%%%%%%%%%%%%%%%%%%%%%%%%%%%%%%%%%%%%%%%%%%%%%%%%%%%%%%%%%
\section{Channel Model and Background} \label{sec: channel model}
%%%%%%%%%%%%%%%%%%%%%%%%%%%%%%%%%%%%%%%%%%%%%%%%%%%%%%%%%%%%%%%%%%%%%%%%%%%%%%%%%%%%%%
%%%%%%%%%%%%%%%%%%%%%%%%%%%%        Subsection            %%%%%%%%%%%%%%%%%%%%%%%%%%%%
%%%%%%%%%%%%%%%%%%%%%%%%%%%%%%%%%%%%%%%%%%%%%%%%%%%%%%%%%%%%%%%%%%%%%%%%%%%%%%%%%%%%%%
\subsection{Channel Model and Soft Demodulation or Equalization}
In this paper, we consider frame-based transmission over a memoryless block fading channel model with constellation $\mathcal{X}$ and channel output's alphabet $\mathcal{Y}$. The channel is characterized by a conditional distribution $p(y|x,c)$ of received symbol $y\in \mathcal{Y}$ given transmitted symbol $x \in \mathcal{X}$ and channel state $c$. In the case of demodulation, we treat the set $\mathcal{X}$ as discrete; while for equalization we view it as the space of vectors of a certain size. In both cases, we will refer to channel input $x$ as symbol. The channel state $c$ is constant within each frame, and it is independently and identically distributed (i.i.d.) across frames according to an unknown distribution $p(c)$. At frame $\tau$, the transmitter sends a packet consisting of $N_\tau$ symbols $x_\tau=\{ x_\tau[i] \}_{i=1}^{N_\tau}$. Given the channel state $c_{\tau}$ and the transmitted symbols, collected in a vector $x_\tau$, the received samples $y_\tau=\{ y_\tau[i] \}_{i=1}^{N_\tau}$ are conditionally independent and each received $i$-th sample is distributed as $y_\tau[i]\sim p(y_\tau[i]|x_\tau[i],c_\tau)$.

A \emph{soft demodulator/equalizer} is a conditional distribution $p(x|y,\phi)$ that maps channel outputs $y\in\mathcal{Y}$ to estimated probabilities for channel input symbol $x\in\mathcal{X}$. The demodulator/equalizer depends on a vector of parameters $\phi$, and is applied separately to each received sample $y[i]$ in a memoryless fashion as $p(x|y_{\tau}[i],\phi)$. The ideal frame-specific parameter vector $\phi_\tau$ for the frame $\tau$ is the one that best approximates the channel conditional distribution $p(x_\tau|y_\tau,c_\tau)$, within its model class, obtained from the Bayes rule as
\begin{equation}
    p(x_\tau|y_\tau,\phi_\tau) \approx p(x_\tau|y_\tau,c_\tau) = \frac{p(y_\tau|x_\tau,c_\tau)p(x_\tau)}{\sum_{x_\tau'\in\mathcal{X}}p(y_\tau|x_\tau',c_\tau)p(x_\tau')},
\end{equation}
where $p(x_\tau)$ is the distribution of the input symbol vector $x_{\tau}$. 
In practice, as we detail below, the demodulator/equalizer is optimized based on pilot symbols. To simplify the terminology, we will also refer to demodulation/equalization as \emph{prediction} henceforth.

%%%%%%%%%%%%%%%%%%%%%%%%%%%%%%%%%%%%%%%%%%%%%%%%%%%%%%%%%%%%%%%%%%%%%%%%%%%%%%%%%%%%%%
%%%%%%%%%%%%%%%%%%%%%%%%%%%%        Subsection            %%%%%%%%%%%%%%%%%%%%%%%%%%%%
%%%%%%%%%%%%%%%%%%%%%%%%%%%%%%%%%%%%%%%%%%%%%%%%%%%%%%%%%%%%%%%%%%%%%%%%%%%%%%%%%%%%%%
\subsection{Conventional Data-Driven Demodulators/Equalizers}
Pilot-aided schemes utilize available pilot symbols to adapt the predictor $p(x|y,\phi)$ to the unknown channel state $c$ in each frame $\tau$. A typical choice for a predictor is a multi-layer neural-network \cite{goodfellow2016deep}. With $L$ layers, given received sample $y$, this class of models produces a vector
\begin{equation}
    a(y|\phi)= W_L \cdot f_{W_{L-1},b_{L-1}}\circ \dots \circ  f_{W_1,b_1} (y) + b_L  ,
\end{equation}
where $\circ$ is the composition operator; the weights $\{W_l\}_{l=1}^L$ and biases $\{b_l\}_{l=1}^L$ define the model parameter vector $\phi:=\{W_l,b_l\}_{l=1}^L$ for a total of $D$ parameters; and the function for the $l$-th layer $f_{W_l,b_l}$ is a linear mapping followed by an entry-wise activation function $h(\cdot)$, i.e., $y_l=f_{W_l,b_l}(y_{l-1})=h(W_l\cdot y_{l-1} + b_l)$ with $y_0=y$. In the last, $L$-th layer, a soft \emph{demodulator} applies the softmax function to vector $a(y|\phi)$, producing the probability distribution
\begin{eqnarray}
\label{eq: demodulator}
    p(x|y,\phi) 
    &=& \big[ \mathrm{softmax} (a(y|\phi)) \big]_x \label{eq: p(x|y,phi) model}\\
    &=& \frac{                           \exp( [ a(y|\phi) ]_x    ) }
             { \sum_{x'\in\mathcal{X}}   \exp( [ a(y|\phi) ]_{x'} ) }, \nonumber
\end{eqnarray}
using $[\cdot]_x$ as the $x$-th element of the vector. In contrast, a soft \emph{equalizer} typically defines the conditional distribution 
\begin{equation}
p(x|y,\phi) = \Normdist (x | a(y|\phi), \beta^{-1} ) , \label{eq: p(x|y,phi) for equalization}
\end{equation}
where the precision $\beta$ is fixed. Throughout this paper, we use $\mathcal{N}(x|\mu,\Sigma)$ to indicate the probability density function of a Gaussian vector with mean $\mu$ and covariance matrix $\Sigma$. 

In each frame $\tau$, \emph{conventional learning} optimizes the model parameters $\phi_\tau$ using $N_\tau^\text{tr}$ i.i.d. pilots $\dataset_\tau^\text{tr}=~\{(y_\tau^\text{tr}[i],x_\tau^\text{tr}[i])\}_{i=1}^{N_\tau^\text{tr}}$ as training data. Optimization of the prediction aims at minimizing the \emph{training log-loss}
\begin{equation}
    \mathcal{L}_{\dataset_\tau^\text{tr}}(\phi_\tau):=-\tfrac{1}{N_\tau^\text{tr}}\sum_{i=1}^{N_\tau^\text{tr}} \log p(x_\tau^\text{tr}[i]|y_\tau^\text{tr}[i],\phi_\tau), \label{eq: L_D_tau^tr(phi_tau)}
\end{equation}
which amounts to the cross entropy for demodulation \eqref{eq: demodulator} and the quadratic prediction loss for equalization \eqref{eq: p(x|y,phi) for equalization}. Minimization of \eqref{eq: L_D_tau^tr(phi_tau)} can be done via \emph{gradient descent} (GD), or stochastic GD (SGD), a variant thereof \cite{Simeone2018Brief}. 

GD updates model parameter vector $\phi_\tau$ for $I$ iterations with learning rate $\eta>0$ starting from an initialization vector $\xi$. Accordingly, the updated parameters $\phi_\tau := \phi^\text{GD}(\dataset_\tau^\text{tr}|\xi)$ are obtained via the iterations
\begin{eqnarray}
    \phi_\tau^{(0)} &=& \xi , \nonumber \\ 
    \forall i=1,\dots,I:\quad  \phi_\tau^{(i)} &\gets& \phi_\tau^{(i-1)} - \eta \nabla_{\phi_\tau^{(i-1)}}  \mathcal{L}_{\dataset_\tau^\text{tr}} (\phi_\tau^{(i-1)}), \nonumber\\
    \phi^\text{GD}(\dataset_\tau^\text{tr}|\xi) &=& \phi_\tau^{(I)} . \label{eq: phi^GD(D|xi,eta,I)}
\end{eqnarray}
The resulting prediction for a test input-output pair $(y_\tau^\text{te}[i],x_\tau^\text{te}[i])$ is given as $p(x_{\tau}^\text{te}[i]|y_{\tau}^\text{te}[i], \phi^\text{GD}(\dataset_{\tau}^\text{tr}|\xi))$.

%%%%%%%%%%%%%%%%%%%%%%%%%%%%%%%%%%%%%%%%%%%%%%%%%%%%%%%%%%%%%%%%%%%%%%%%%%%%%%%%%%%%%%
%%%%%%%%%%%%%%%%%%%%%%%%%%%%        Subsection            %%%%%%%%%%%%%%%%%%%%%%%%%%%%
%%%%%%%%%%%%%%%%%%%%%%%%%%%%%%%%%%%%%%%%%%%%%%%%%%%%%%%%%%%%%%%%%%%%%%%%%%%%%%%%%%%%%%%%%%%%%%%%%%%%%%%%%%%%%%%%%%%%%%%%%%%%%%%%%%%%%%%%%%%%%%%
\subsection{Frequentist Meta-Learning} \label{subsec: Frequentist Meta-Learning}

The most prominent shortcoming of conventional learning is its potentially high sample complexity, which translates into the need for a large number of pilots, $N_\tau^\text{tr}$, per frame. Meta-learning addresses this issue by transferring knowledge acquired over previous frames. Specifically, frequentist meta-learning, as proposed in \cite{park2021fewpilots}, treats the initialization vector $\xi$ in \eqref{eq: phi^GD(D|xi,eta,I)} as a hyperparameter vector to be optimized based on the availability of pilots from $t$ previous transmission frames.

As a preliminary step, we decompose the available pilots from each frame $\tau$ into a disjoint training set $\dataset_\tau^\text{tr}$ and test set $\dataset_\tau^\text{te}$ as $\dataset_\tau = \{\dataset_\tau^\text{tr},  \dataset_\tau^\text{te}\}$. Furthermore, the data sets for all previous $t$ frames are stacked as $\dataset_{1:t}=\{\dataset_\tau\}_{\tau=1}^t$, and similarly for $\dataset_{1:t}^\text{te}=\{\dataset^\text{te}_\tau\}_{\tau=1}^t$, having a total of $N_{1:t}^\text{te}=\sum_{\tau=1}^tN_\tau^\text{te}$ samples. Meta-learning has two phases: meta-training and meta-testing. These are defined next by following the frequentist meta-learning strategy of \cite{park2021fewpilots}.

\textit{Meta-training} tackles the bi-level optimization problem
\begin{subequations}
    \label{eq: freq meta-learning as optim problem}
    \begin{eqnarray}
        \min\limits_{\xi} &\;& \negspaceF \tfrac{1}{N_{1:t}^\text{te}} \sum_{\tau=1}^t N_\tau^\text{te} \mathcal{L}_{\dataset_\tau^\text{te}}\big(\phi_\tau(\dataset_\tau^\text{tr}|\xi)\big) \label{eq: freq outer} \\
        \textrm{s.t.}    &\;& \negspaceF \phi_\tau(\dataset_\tau^\text{tr}|\xi) = \argmin_{\phi(\xi)}  \mathcal{L}_{\dataset_\tau^\text{tr}}(\phi), \quad \tau=1,\dots,t. \label{eq: freq inner}
    \end{eqnarray}
\end{subequations}The notation $\phi(\xi)$ in \eqref{eq: freq inner} indicates the dependence of the optimizer on the initialization vector $\xi$.
By \eqref{eq: freq meta-learning as optim problem}, the goal of frequentist meta-training is to find a hyperparameter vector $\xi$ such that for any frame $\tau$, the optimized model parameter vector $\phi_\tau(\dataset_\tau^\text{tr}|\xi)$ fits well the test data set  $\dataset_\tau^\text{te}$.

Problem \eqref{eq: freq meta-learning as optim problem} is addressed via a nested loop optimization involving SGD-based inner updates and SGD-based outer updates, which are also referred as meta-iterations. The inner loop tackles the inner optimization \eqref{eq: freq inner} in a per-frame manner via \eqref{eq: phi^GD(D|xi,eta,I)} for a randomly selected subset $\mathcal{T}\subset\{1,\dots,t\}$ of frames, which are redrawn independently at each meta-iteration. The outer loop addresses the outer optimization \eqref{eq: freq outer} via an SGD step of the meta-loss with learning-rate $\kappa>0$, i.e.,
\begin{equation}
    \xi \gets \xi - \kappa \frac{1}{N_\mathcal{T}^\text{te}} \sum_{\tau\in \mathcal{T}} N_\tau^\text{te} \nabla_\xi \mathcal{L}_{\dataset_\tau^\text{te}} \big(\phi^\text{GD} (\dataset_\tau^\text{tr}|\xi) \big) , \label{eq: xi meta update freq}
\end{equation}
based on data from the batch $\mathcal{T}$ of selected frames, and using the notation $N_\mathcal{T}^{\text{te}}=\sum_{\tau\in\mathcal{T}} N_\tau^{\text{te}}$ for the total samples within the batch of selected frames. Meta-training updates the initialization vector $\xi$ across multiple meta-iterations. When meeting some stopping criterion, here determined by a predefined number of meta-iterations $I_\text{meta}$, meta-training stops, and the hyperparameter vector $\xi$ is stored to be used for future learning tasks.

Upon deployment, i.e., during \textit{meta-testing}, the meta-test frames also include pilots and data as the meta-training frames. Accordingly, each meta-test device loads the hyperparameter vector $\xi$ for initialization, and produces the adapted model parameter vector $\phi_*=\phi^\text{GD}(\mathcal{D}_*^\text{tr}|\xi)$ as in \eqref{eq: phi^GD(D|xi,eta,I)} using $N_*^\text{tr}$ pilots symbols $\mathcal{D}_*^\text{tr} = \{ (y_*^\text{tr}[i], x_*^\text{tr}[i]) \}_{i=1}^{N_*^\text{tr}}$. Then, it applies the learned model to the payload data symbols $\{y_*^\text{te}[i]\}_{i=1}^{N_*^\text{te}}$ to carry out demodulation or equalization
\begin{equation}
    p(x_*^\text{te}[i]|y_*^\text{te}[i],\phi_*).  \label{eq:freqpred}
\end{equation}

%%%%%%%%%%%%%%%%%%%%%%%%%%%%%%%%%%%%%%%%%%%%%%%%%%%%%%%%%%%%%%%%%%%%%%%%%%%%%%%%%%%%%%%%%%%%%%%%%%%%%%%%%%
%%%%%%%%%%%%%%%%%%%%%%%%%%%%%%%%%%%%%%          Section             %%%%%%%%%%%%%%%%%%%%%%%%%%%%%%%%%%%%%%
%%%%%%%%%%%%%%%%%%%%%%%%%%%%%%%%%%%%%%%%%%%%%%%%%%%%%%%%%%%%%%%%%%%%%%%%%%%%%%%%%%%%%%%%%%%%%%%%%%%%%%%%%%
\section{The Bayesian Framework} \label{sec: Bayesian VI ML}

%%%%%%%%%%%%%%%%%%%%%%%%%%%%%%%%%%%%%%%%%%%%%%%%%%%%%%%%%%%%%%%%%%%%%%%%%%%%%%%%%%%%%%
%%%%%%%%%%%%%%%%%%%%%%%%%%%%        Subsection            %%%%%%%%%%%%%%%%%%%%%%%%%%%%
%%%%%%%%%%%%%%%%%%%%%%%%%%%%%%%%%%%%%%%%%%%%%%%%%%%%%%%%%%%%%%%%%%%%%%%%%%%%%%%%%%%%%%
\subsection{Bayesian Learning} \label{subsec: Bayesian Learning}

\emph{Bayesian learning} treats the model parameter vector $\phi_\tau$ for some frame $\tau$ as a random vector, rather than as a deterministic optimization variable as in frequentist learning framework. As illustrated in Fig.~\ref{Fig: Evolution of network, determinitstic to Bayesian amortized VI}, instead of producing a single demodulator parameters $\phi_\tau=\phi^\text{GD}(\dataset_\tau^\text{tr}|\xi)$ as in \eqref{eq: phi^GD(D|xi,eta,I)}, Bayesian learning produces a distribution $p(\phi_\tau | \dataset_\tau^\text{tr},\xi)$ over the space of the demodulator parameters $\phi_\tau$. This distribution is computed based on training data $\dataset_\tau^\text{tr}$ and on \emph{predetermined} prior distribution $p(\phi_\tau | \xi)$, which depends in turn on the hyperparameter vector $\xi$, also fixed \emph{a priori}.

Having obtained the distribution $p(\phi_\tau | \dataset_\tau^\text{tr},\xi)$, the \emph{ensemble prediction} of a test point $(y_\tau^\text{te}[i],x_\tau^\text{te}[i])$ is given by the ensemble average of the predictions $p(x_\tau^\text{te}[i]|y_\tau^\text{te}[i],\phi_\tau)$ with random vector $\phi_\tau$ having distribution $p(\phi_\tau | \dataset_\tau^\text{tr},\xi)$, i.e.,
\begin{equation}
    p \big(x_{\tau}^\text{te}[i] \big| y_{\tau}^\text{te}[i],\dataset_{\tau}^\text{tr},\xi \big) = \E_{p(\phi_\tau | \dataset_\tau^\text{tr},\xi)} \big[ p \big(x_{\tau}^\text{te}[i] \big| y_{\tau}^\text{te}[i], \phi_\tau \big) \big]. \label{eq: p(x|y,Dtr,xi)=E_p[p(x|y,phi))]}
\end{equation}
The frequentist prediction \eqref{eq:freqpred}, reviewed in the previous section, can be viewed as a special case in which one is limited to the choice $p(\phi_\tau | \dataset_\tau^\text{tr}, \xi) = \delta(\phi_\tau - \phi^\text{GD}(\dataset_\tau^\text{tr}|\xi))$, with $\delta(\cdot)$ indicating the Dirac Delta. With this choice, the distribution $p(\phi_\tau | \dataset_\tau^\text{tr},\xi)$ is concentrated at one point, namely the GD solution \eqref{eq: phi^GD(D|xi,eta,I)}. The frequentist approach is therefore inherently limited in its capacity to express uncertainty on the model parameters due to limited data.

Ideally, the distribution $p(\phi_\tau | \dataset_\tau^\text{tr},\xi)$ should be obtained as the posterior distribution
\begin{equation}
    p(\phi_\tau | \dataset_\tau^\text{tr},\xi) \propto p(\phi_\tau | \xi) p(\dataset_\tau^\text{tr} | \phi_\tau), \label{eq: Bayes Theorem of p(phi|D,xi) }
\end{equation}
where $p(\dataset_\tau^\text{tr} | \phi_\tau) = \prod_{i=1}^{N_\tau^\text{tr}}p(x_\tau^\text{tr}[i]|y_\tau^\text{tr}[i],\phi_\tau) $ is the likelihood function for the training data. However, computing the posterior $p(\phi_\tau | \dataset_\tau^\text{tr},\xi)$ in \eqref{eq: Bayes Theorem of p(phi|D,xi) } is generally intractable for high dimensional vector $\phi_\tau$. 

To address this challenge, we follow VI and introduce a \emph{variational distribution} approximation
\begin{equation}
    q(\phi_\tau|\varphi_\tau) \approx p(\phi_\tau | \dataset_\tau^\text{tr}, \xi), \label{eq: variational inference q approx posterior}
\end{equation}
which depends on a variational parameter vector $\varphi_\tau$. A typical choice is given by the Gaussian mean-field approximation \cite{angelino2016patterns} which can be expressed as
\begin{equation}
    q(\phi_\tau|\varphi_\tau) = \Normdist(\phi_\tau | \nu_\tau,\Diag(\exp(2\varrho_\tau))), \label{eq: Gaussian prior q}
\end{equation}
with variational parameter vector $\varphi_\tau = [ \nu_\tau^\top , \varrho_\tau^\top ]^\top$, and the exponent function is applied element-wise. The variational parameter vector includes the mean vector $\nu_\tau\in\Rset^D$ and the vector of the logarithm of the standard deviations $\varrho_\tau\in\Rset^D$ for the Gaussian random vector $\phi_\tau$. Note that vector $\varrho_\tau$ models  uncertainty in the model parameter space.

To describe VI, we will use the Kullback-Liebler (KL) divergence $\KL(p(z)||q(z))$ \cite{cover1991information}, which is a measure of the distance between two distributions $p(z)$ and $q(z)$. It is defined as the average of the log-likelihood ratio $\log(p(z)/q(z))$ as 
\begin{equation} 
    \KL(p(z)||q(z))= \E_{p(z)}\Big[ \log \Big( \frac{p(z)}{q(z)} \Big) \Big]. \label{eq: KL definition}
\end{equation}
VI-based Bayesian learning prescribes that the variational parameter vectors $\phi_\tau$ be obtained via the minimization of the KL divergence $\KL \adjpavv{q(\phi_\tau|\varphi)}{p(\phi_\tau | \mathcal{D}_\tau^\text{tr}, \xi)}$ between the variational distribution $q(\phi_\tau|\varphi)$ and the posterior distribution $p(\phi_\tau | \mathcal{D}_\tau^\text{tr}, \xi)$. This problem can be equivalently formulated as the minimization  \cite{angelino2016patterns,Simeone2018Brief} 
\begin{equation}
   \varphi_\tau = \argmin_{\varphi}F_{\dataset_\tau^\text{tr}}(\varphi|\xi), \label{eq: varphi_tau = argmin F_D_tau^tr}
\end{equation}
where the \emph{variational free energy} \cite{jose2021free} is defined as
\begin{eqnarray}
    F_{\dataset_\tau^\text{tr}}(\varphi_\tau|\xi)
    \negspaceF&=&\negspaceF  N_\tau^\text{tr} \E_{q(\phi_\tau|\varphi_\tau)}  [\mathcal{L}_{\dataset_\tau^\text{tr}}(\phi_\tau)]  +\KL(q(\phi_\tau|\varphi_\tau)||p(\phi_\tau|\xi)) \nonumber\\
    \negspaceF&=&\negspaceF  N_\tau^\text{tr} L_{\dataset_\tau^\text{tr}}(\varphi_\tau) +\KL(q(\phi_\tau|\varphi_\tau)||p(\phi_\tau|\xi)).
    \label{eq: F_D_tau(varphi,xi) def}
\end{eqnarray} 
In \eqref{eq: F_D_tau(varphi,xi) def}, we have defined as $L_{\dataset_\tau^\text{tr}}(\varphi_\tau)$ the expectation of loss function $\mathcal{L}_{\dataset_\tau^\text{tr}}(\phi_\tau)$ \eqref{eq: L_D_tau^tr(phi_tau)} over variational distribution $q(\phi_\tau|\varphi_\tau)$, i.e.,
 \begin{eqnarray}
     L_{\dataset_\tau^\text{tr}}(\varphi_\tau) = \E_{q(\phi_\tau|\varphi_\tau)}  [\mathcal{L}_{\dataset_\tau^\text{tr}}(\phi_\tau)]. \label{eq: def L_D = E[L_D]}
 \end{eqnarray}

 In \eqref{eq: F_D_tau(varphi,xi) def}, the second summand  is a regularizer that restricts the variational distribution to be close to the prior distribution. Note that, if the variational distribution has ability to express the posterior distribution in \eqref{eq: Bayes Theorem of p(phi|D,xi) }, the minimizer of the problem \eqref{eq: varphi_tau = argmin F_D_tau^tr} becomes the Bayesian posterior $p(\phi_\tau|\dataset_\tau^\text{tr},\xi)$, since the KL divergence $\KL \adjpavv{q(\phi_\tau|\varphi)}{p(\phi_\tau | \mathcal{D}_\tau^\text{tr}, \xi)}$ is minimized (and it equals zero) when the two distributions are the same. 
 
A typical choice for the prior distribution $p(\phi_\tau|\xi)$ is the Gaussian distribution. In this case, we have
\begin{equation}
    p(\phi_\tau | \xi)   = \Normdist(\phi_\tau  | \nu ,\Diag(\exp(2\varrho ))) , \label{eq: Gaussian prior p(phi_tau|xi)}
\end{equation}
which is defined by the hyperparameter vector $\xi = [ \nu^\top , \varrho^\top]^\top$, where $\nu\in\Rset^D$ and $\varrho\in\Rset^D$ stand for the mean and logarithm of the standard deviation vector of the Gaussian random vector $\phi_\tau$.

Assuming the Gaussian variational distribution in \eqref{eq: Gaussian prior q} and the Gaussian prior \eqref{eq: Gaussian prior p(phi_tau|xi)}, the regularizer term in \eqref{eq: F_D_tau(varphi,xi) def} can be computed in closed-form as \\$\KL(q(\phi_\tau | \varphi_\tau)||p(\phi_\tau|\xi))=$
\begin{equation*}
    \tfrac{1}{2}\sum_{d=1}^D\adjpa{
         2(\varrho[d]-\varrho_\tau[d])+\frac{\exp(2\varrho_\tau[d])+(\nu_\tau[d]-\nu[d])^2}{\exp(2\varrho[d])}-1 
        } , \label{eq: KL(q p) both gaussians}
\end{equation*}
which is a differentiable function for $\varphi_\tau$. 

With these choices of variational posterior and prior, problem \eqref{eq: varphi_tau = argmin F_D_tau^tr} can be addressed via gradient-descent methods by using the reparametrization trick \cite{kingma2013autoencoding}. This is done by writing the random model parameter vector $\phi_\tau \sim q(\phi_\tau |\varphi_\tau)$ as $\phi_\tau~=~\nu_\tau~+~\exp{(\varrho_\tau)} \odot \rv{e}$, with random vector $\rv{e} \sim \mathcal{N}(0,I_D)$ and $\odot$ being the element-wise multiplication. An estimate of the gradient of the objective \eqref{eq: def L_D = E[L_D]} using the reparametrization trick is done with the aid of $R$ drawn independently samples of the standard normal Gaussian random vector $\rv{e}$, and differentiating the resulting empirical estimate of \eqref{eq: def L_D = E[L_D]}. 

%\vspace{-0.2cm}
\begin{algorithm}%[H]
    \caption{\texttt{Reparametrization Trick } \cite{kingma2013autoencoding}}
    \label{alg: reparm trick}
    \KwInputs{      $\mathcal{G}(\cdot)$ = a function over vector $\phi_\tau$ \\
                    $\varphi_\tau$ = variational parameter}
    \KwParameters{  $R$ = ensemble size}
    \KwOutput{      $\hat{G}(\varphi_\tau)$ = approximation of  $\E_{q(\phi_\tau|\varphi_\tau)}[\mathcal{G}(\phi_\tau)] $}
    \BlankLine
    
    \For{$r= 1,\dots,R$}{
        Draw $\rv{e}_{\tau,r} \sim \Normdist(0,I_D)$\;
        $\phi_{\tau,r}(\varphi_\tau,\rv{e}_{\tau,r}) \gets \nu_\tau + \exp{(\varrho_\tau)} \odot \rv{e}_{\tau,r}$\; \label{algline: reparam trick liner transform}
    }
    \Return $\hat{G}(\varphi_\tau) \gets \tfrac{1}{R}\sum_{r=1}^R \mathcal{G}\big( \phi_{\tau,r}(\varphi_\tau,\rv{e}_{\tau,r}) \big)$ 
\end{algorithm}

Specifically, we estimate the free energy in \eqref{eq: F_D_tau(varphi,xi) def} by replacing the training loss ${L}_{\mathcal{D}_\tau^\text{tr}}(\varphi_\tau)$ with the empirical estimate
\begin{align}
    \hat{L}_{\dataset_\tau^\text{tr}}(\varphi_\tau) = \frac{1}{R} \sum_{r=1}^R \mathcal{L}_{\mathcal{D}_\tau^\text{tr}}\big(\nu_\tau + \exp{(\varrho_\tau)} \odot \rv{e}_{\tau,r}\big),
\end{align}
obtained by drawing samples $\rv{e}_{\tau,r} \sim \mathcal{N}(0, I_D)$ for $r~=~1,2,\dots,R$. This yields the estimated free energy
\begin{equation}
    \hat{F}_{\dataset_\tau^\text{tr}}(\varphi_\tau|\xi)
    =  N_\tau^\text{tr} \hat{L}_{\dataset_\tau^\text{tr}}(\varphi_\tau) +\KL(q(\phi_\tau|\varphi)||p(\phi_\tau|\xi)).  \label{eq: F_hat_D_tau(varphi,xi) def}
\end{equation}  
This is a special case of Algorithm~\ref{alg: reparm trick} with input $\mathcal{G}(\phi_\tau)=\mathcal{L}_{\dataset_\tau^\text{tr}}(\phi_\tau)$. The function \eqref{eq: F_hat_D_tau(varphi,xi) def} can be directly differentiated and used in SGD updates.

Once the variational parameter $\varphi_\tau$ is inferred using Bayesian training, ensemble prediction for a payload data symbol $(y_\tau^\text{te}[i],x_\tau^\text{te}[i])$ can be obtained via \eqref{eq: p(x|y,Dtr,xi)=E_p[p(x|y,phi))]} by replacing $p(\phi_\tau | \dataset_\tau^\text{tr},\xi)$ with $q(\phi_\tau|\varphi_\tau)$ to yield the ensemble predictor
\begin{equation}
    p(x_{\tau}^\text{te}[i]|y_{\tau}^\text{te}[i],\varphi_\tau) = \E_{q(\phi_\tau|\varphi_\tau)} \big[ p(x_{\tau}^\text{te}[i]|y_{\tau}^\text{te}[i], \phi_\tau) \big].
    \label{eq:bayesian_ensemble_prediction}
\end{equation}
Practically, it uses Monte Carlo sampling with $R$ model vectors, producing the approximated soft predictor $\hat{p}(x_{\tau}^\text{te}[i]|y_{\tau}^\text{te}[i],\varphi_\tau)$ via Algorithm~\ref{alg: reparm trick} with $\mathcal{G}(\phi_\tau)=p(x_{\tau}^\text{te}[i]|y_{\tau}^\text{te}[i],\phi_\tau)$.

%%%%%%%%%%%%%%%%%%%%%%%%%%%%%%%%%%%%%%%%%%%%%%%%%%%%%%%%%%%%%%%%%%%%%%%%%%%%%%%%%%%%%%
%%%%%%%%%%%%%%%%%%%%%%%%%%%%        Subsection            %%%%%%%%%%%%%%%%%%%%%%%%%%%%
%%%%%%%%%%%%%%%%%%%%%%%%%%%%%%%%%%%%%%%%%%%%%%%%%%%%%%%%%%%%%%%%%%%%%%%%%%%%%%%%%%%%%%
\subsection{Bayesian Meta-Learning}\label{sec: Bayesian Meta-Learning}

While conventional Bayesian learning assumes that the random model parameter vector $\phi_\tau$ has a fixed prior distribution $p(\phi_\tau|\xi)$ parametrized by a predefined hyperparameter vector $\xi$, Bayesian meta-learning leverages the stronger assumption that there is a shared prior distribution $p(\phi_\tau|\xi)$ across all frames that can be optimized through a hyperparameter vector $\xi$.

\begin{figure}
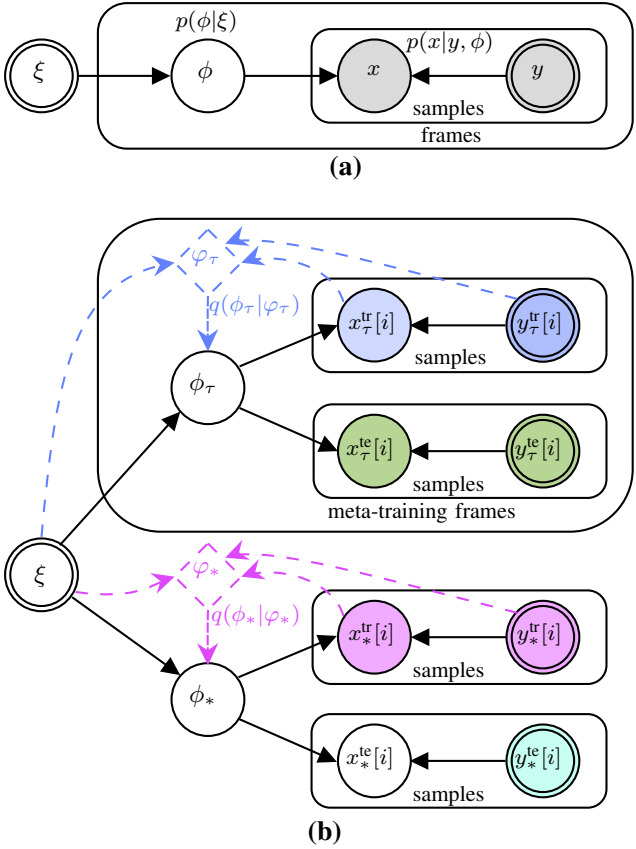
%[ht!]
    \centering
    \includestandalone[width=8.5cm]{fig_tikz_prob_graphical_model}%     without .tex extension
    \caption{Probabilistic graphical model (Bayesian network) \cite{koller2009probabilistic} for Bayesian meta-learning. Circles represent random variables; double-lined circles represent deterministic variables or (hyper)parameters; shaded circles represent observations; dashed diamonds represent variational parameter vectors; and plaques indicate multiple instances (the outer plaques represent frames, whereas the inner represent multiple sample, e.g., symbols across time): (a) High level representation, assuming a prior $p(\phi|\xi)$ and predictor $p(x|y,\phi)$; (b) Model using the train/test splits, with variational inference $q(\phi_\tau|\varphi_\tau) \approx p(\phi_\tau | \dataset_\tau^\text{tr}, \xi)$ indicated as dashed arrows.}
    \label{fig: Probabilistic Graphical Models}
\end{figure}

In this section, we formulate Bayesian meta-learning by following \emph{empirical Bayes} \cite{grant2018recasting}, with the aim of selecting a distribution $p(\phi_\tau|\xi)$ that provides a useful prior for the design of the predictor on new frames. Mathematically, Bayesian meta-training optimizes over the hyperparameter vector $\xi$ by addressing the bi-level problem
 \begin{subequations}
     \label{eq: Bayesian meta-learning as optim problem}
     \begin{eqnarray}
         \negspaceF\negspaceF \min\limits_{\xi}  && \negspaceF \tfrac{1}{N_{1:t}^\text{te}} \sum_{\tau=1}^t N_\tau^\text{te} \E_{q(\phi_\tau|\varphi_\tau(\dataset_\tau^\text{tr}|\xi))} \big[ \mathcal{L}_{\dataset_\tau^\text{te}}(\phi_\tau) \big] \label{eq: passive bayesian outer} \\
         \negspaceF\negspaceF \textrm{s.t.}      && \negspaceF \varphi_\tau(\dataset_\tau^\text{tr}|\xi) = \argmin_{\varphi}  F_{\dataset_\tau^\text{tr}}(\varphi|\xi), \tau=1,\dots,t. \label{eq: passive bayesian inner}
     \end{eqnarray}
 \end{subequations}
Problem \eqref{eq: Bayesian meta-learning as optim problem} chooses the hyperparameter vector $\xi$ that minimizes the average test loss on the meta-training frames $\tau \in\{1,\dots,t\}$ that is obtained with the variational posterior via \eqref{eq: varphi_tau = argmin F_D_tau^tr}. The subproblems in \eqref{eq: passive bayesian inner} correspond to Bayesian learning applied separately to each frame as explained in Section \ref{subsec: Bayesian Learning}. An illustration of all the quantities involved in problem \eqref{eq: Bayesian meta-learning as optim problem} can be found in Fig.~\ref{fig: Probabilistic Graphical Models} by using the formalism of Bayesian networks \cite{koller2009probabilistic}.

To address problem \eqref{eq: Bayesian meta-learning as optim problem} in a tractable manner, we apply the  reparametrization trick for both outer \eqref{eq: passive bayesian outer} and inner optimization \eqref{eq: passive bayesian inner} by following the same steps described in Section \ref{subsec: Bayesian Learning}.  Details on the optimization can be found in Algorithm~\ref{alg: bayesian maml meta-training}. In short, the inner loop updates the frame-specific variational parameters $\varphi_\tau$ by minimizing the approximated free energy  \eqref{eq: F_hat_D_tau(varphi,xi) def} separately for each frame $\tau$ within a mini-batch $\mathcal{T}$  via GD (dashed blue line in Fig.~\ref{fig: Probabilistic Graphical Models}b). Following \cite{nguyen2020uncertainty,ravi2018amortized}, the prior's parameter vector $\xi$ plays two roles in the inner loop, namely \emph{(i)} as the initialization for the inner GD update in Algorithm~\ref{alg: bayesian maml meta-training} line~\ref{algline: phi_tau update}; and \emph{(ii)} as the regularizer for the same update via the prior $p(\phi_\tau|\xi)$. 
The outer optimization \eqref{eq: passive bayesian outer} is addressed via SGD to minimize the average log-likelihood for test set using Algorithm~\ref{alg: reparm trick} with $\mathcal{G}(\phi_\tau)=\mathcal{L}_{\dataset_\tau^\text{te}}(\phi_\tau)$, shown as dashed green line in Fig.~\ref{fig: Probabilistic Graphical Models}b. 

%\vspace{-0.2cm}
\begin{algorithm}%[H]
    \caption{\texttt{Bayesian Meta-Training}}
    \label{alg: bayesian maml meta-training}
    \KwInputs{      $\dataset_{1:t}$ = labeled data sets of $t$ meta-training frames}
    \KwParameters{  $B$ = number of frames per meta-update batch\\
                    $I$ = number of inner update steps\\
                    $\eta,\kappa$ = inner/outer updates learning rates \\
                    }
    \KwOutput{      $\xi$ = learned hyperparameter vector}
    \BlankLine
    initialize $\xi$\;
    \While{meta-learning not done}{
        $\mathcal{T} \gets$ random batch of $B$ frames \;
        \For{$\tau\in \mathcal{T}$}{
            randomly divide $\dataset_\tau = \{ \dataset_\tau^\text{tr},\dataset_\tau^\text{te}\}$\; \label{algline: bayesian dataset to tr and te}
            \algcomment{frame-specific update}\\
            $ \varphi_\tau^{(0)} \gets \xi$\;
            \For{$i=1,2,\dots,I$ inner update steps}{
                $ \varphi_\tau^{(i)} \!\! \gets \! \varphi_\tau^{(i-1)} \!\!\! - \! \tfrac{\eta}{N_\tau^\text{tr}} \nabla_{\varphi_\tau}  \hat{F}_{\dataset_\tau^\text{tr}} \big( \varphi_\tau^{(i-1)} \big| \xi\big)$  using \eqref{eq: F_hat_D_tau(varphi,xi) def} \label{algline: phi_tau update}
            }
           $\varphi^\text{GD}(\dataset_\tau^\text{tr}|\xi) \gets \varphi_\tau^{(I)}$\;
        }
        \algcomment{meta-update}\\
        $\xi \gets \xi - \kappa\frac{1}{N_\mathcal{T}^\text{te}} \sum_{t\in\mathcal{T}} N_\tau^\text{te} \nabla_{\xi} \hat{L}_{\dataset_\tau^\text{te}}\big(\varphi^\text{GD}(\dataset_\tau^\text{tr}|\xi)\big)$
    }
    \Return $\xi$
\end{algorithm}
 
\begin{figure}%[!ht]
    \centering
    \includegraphics[page=1,trim=0cm 0cm 5.5cm 0cm, clip, width=9cm]{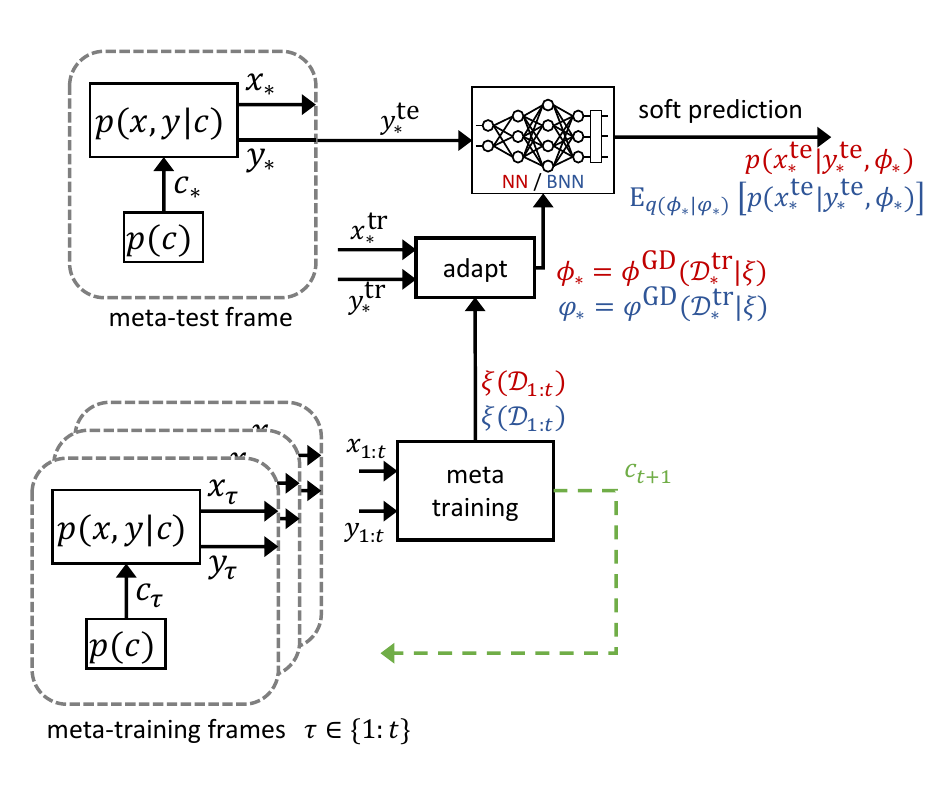}
    \caption{Bayesian meta-learning (blue) as compared to frequentist meta-learning (red). The frequentist predictor uses a single predictor, depicted as a neural network (NN), whereas Bayesian meta-learning uses an ensemble of predictors, e.g., a Bayesian NN (BNN). The dashed line represents the operation of the active meta-learning introduced in Section~\ref{sec: Bayesian Active Meta-Learning}. The data for each frame is generated by following the distribution $p(x,y|c)=p(x)p(y|x,c)$, with input distribution $p(x)$ and conditional distribution $p(y|x,c)$ for channel state $c$.}
    \label{fig: freq meta-learning vs bayesian meta-learning}
\end{figure}

After obtaining meta-trained hyperparameter $\xi$, meta-testing takes place, starting with the adaptation of the variational parameter $\varphi_*(\dataset_*^\text{tr}|\xi)$ via \eqref{eq: passive bayesian inner} using the available pilot data $\dataset_*^\text{tr}$ at the current frame, to obtain ensemble prediction 
\begin{equation}
    p(x_*^\text{te}[i]|y_*^\text{te}[i],\varphi_*) = \E_{q(\phi_*|\varphi_*)} \big[ p(x_*^\text{te}[i]|y_*^\text{te}[i], \phi_*) \big],
\end{equation}
as done in \eqref{eq:bayesian_ensemble_prediction}. Bayesian meta-learning is illustrated comparatively to meta-learning in Fig.~\ref{fig: freq meta-learning vs bayesian meta-learning}.

%%%%%%%%%%%%%%%%%%%%%%%%%%%%%%%%%%%%%%%%%%%%%%%%%%%%%%%%%%%%%%%%%%%%%%%%%%%%%%%%%%%%%%
%%%%%%%%%%%%%%%%%%%%%%%%%%%%        Subsection            %%%%%%%%%%%%%%%%%%%%%%%%%%%%
%%%%%%%%%%%%%%%%%%%%%%%%%%%%%%%%%%%%%%%%%%%%%%%%%%%%%%%%%%%%%%%%%%%%%%%%%%%%%%%%%%%%%%
\subsection{Computational Complexity}\label{sec: Computational Complexity}

We now briefly elaborate on the complexity of meta-learning by analyzing the complexity of meta-training and of meta-testing. To this end, let us define as $C$ the complexity of obtaining the probability $p(x|y,\phi)$ for a data sample $(y,x)$. This baseline complexity depends on the model dimensionality, and it accounts for the amount of time needed to carry out the forward pass on the neural network implementing the model $p(x|y,\phi)$. Accordingly, as seen in Table~\ref{tab: Computational complexity order of frequentist and Bayesian meta-learning}, the per-data point complexity of meta-testing equals $C$ for frequentist learning, and $C R^\text{te}$ for Bayesian learning, where $R^\text{te}$ is the size of the ensemble used for inference.

The complexity of computing the first-order gradient via backpropagation per-sample is given by $G_1 C$, with $G_1$ being a constant in the range between $2$ and $5$ \cite{simeone2022machine,griewank1993some}. Furthermore, computing the Hessian-vector product (HVP) has a complexity of the order $G_2 G_1 C$, where constant $G_2$ is also between $2$ to $5$ \cite[Appendix A]{park2021fewpilots},\cite[Appendix C]{rajeswaran2019meta}. Assume that all tasks have data sets of equal size, i.e., $N_\tau^\text{tr}=N^\text{tr}$ and $N_\tau^\text{te}=N^\text{te}$ for any task $\tau$. Therefore, for each meta-training iteration, for a batch of $B$ tasks with $I$ local updates, the complexity of the frequentist meta-update \eqref{eq: xi meta update freq} is of the order 
\begin{equation}
    B 
    \Big(
    \underbrace{I N^\text{tr} G_1   C  }_\text{frame-specific update}   + 
    \underbrace{I N^\text{tr} G_2 G_1 C  }_\text{HVPs in meta-update} + 
    \underbrace{  N^\text{te} G_1   C  }_\text{gradient in meta-update}
    \Big) .
\end{equation}

For Bayesian meta-learning, the complexity increases linearly with the training ensemble size that is used for estimating the loss functions in \eqref{eq: passive bayesian outer} and \eqref{eq: passive bayesian inner}. Note that the impact of the size $R^\text{tr}$ of the training ensemble used for meta-training is different from the size $R^\text{te}$ used for inference, as the first determines the variance of the stochastic loss functions, while the latter determines the quality of Bayesian prediction (see, e.g., \cite{zecchin2022robust} and references therein). Ignoring the constant cost of differentiating the KL term in the free energy and for sampling from the Gaussian distribution, the complexity analysis is summarized in Table~\ref{tab: Computational complexity order of frequentist and Bayesian meta-learning}.

\begin{table}[ht]
    \centering
    \caption{Computational complexity of frequentist and Bayesian meta-learning. (See text in Sec.~\ref{sec: Computational Complexity} for details)}
    \label{tab: Computational complexity order of frequentist and Bayesian meta-learning}
    %\begin{tabular}{ l c c }
    \begin{tabular}{ p{1.1cm} p{1.55cm} p{4.4cm} }
    \toprule
    %    & \vtop{\hbox{\strut inference}\hbox{\strut [per-test sample]}}
    %    & \vtop{\hbox{\strut meta-training}\hbox{\strut [per-meta-iteration]}}
        & inference [per-test sample]
        & meta-training [per-meta-iteration]
      \\ [0.5ex] \midrule\midrule
        frequentist 
        & $C$               
        & $ B G_1 C \big(I N^\text{tr} (G_2 +1) + N^\text{te}\big)$
      \\ [0.5ex] 
        Bayesian
        & $C   R^\text{te}$ 
        & $ B G_1 C \big(I N^\text{tr} (G_2 +1)R^\text{tr} + N^\text{te}R^\text{te}\big) $
      \\ %[0.5ex] 
      \bottomrule
    \end{tabular}
    \vspace{0.2cm}
\end{table}

%%%%%%%%%%%%%%%%%%%%%%%%%%%%%%%%%%%%%%%%%%%%%%%%%%%%%%%%%%%%%%%%%%%%%%%%%%%%%%%%%%%%%%%%%%%%%%%%%%%%%%%%%%
%%%%%%%%%%%%%%%%%%%%%%%%%%%%%%%%%%%%%%          Section             %%%%%%%%%%%%%%%%%%%%%%%%%%%%%%%%%%%%%%
%%%%%%%%%%%%%%%%%%%%%%%%%%%%%%%%%%%%%%%%%%%%%%%%%%%%%%%%%%%%%%%%%%%%%%%%%%%%%%%%%%%%%%%%%%%%%%%%%%%%%%%%%%
\section{Bayesian Active Meta-Learning}\label{sec: Bayesian Active Meta-Learning}

In the previous sections, we have considered a passive meta-learning setting in which the meta-learner is given a number of meta-training data sets, each corresponding to a different channel state $c$. In this section, we study the situation in which the meta-learner has access to a simulator that can be used to generate random data sets for any channel state $c$ via the channel $p(y|x,c)$. The problem of interest is to minimize the use of the simulator by actively selecting the channels $\{c_\tau\}$ for which meta-training data is generated. To this end, we devise a sequential approach, whereby the meta-learner optimizes the next channel state $c_{t+1}$, given all $t$ meta-training data sets of frames $\tau=1,\dots,t$.

At the core of the proposed active meta-learning strategy, are mechanisms used by the meta-learner to discover model parameter vectors $\phi$ that have been underexplored so far, and to relate model parameter vector $\phi$ to a channel state.

%%%%%%%%%%%%%%%%%%%%%%%%%%%%%%%%%%%%%%%%%%%%%%%%%%%%%%%%%%%%%%%%%%%%%%%%%%%%%%%%%%%%%%
%%%%%%%%%%%%%%%%%%%%%%%%%%%%        Subsection            %%%%%%%%%%%%%%%%%%%%%%%%%%%%
%%%%%%%%%%%%%%%%%%%%%%%%%%%%%%%%%%%%%%%%%%%%%%%%%%%%%%%%%%%%%%%%%%%%%%%%%%%%%%%%%%%%%%
\subsection{Active Selection of Channel States}
\label{subsec:active_selec_of_channel_states}

After having collected $t$ meta-training data sets $\dataset_{1:t}=\{\dataset_\tau\}_{\tau=1}^t$, the proposed active meta-learning scheme selects the next channel state, $c_{t+1}$, to use for the generation of the $(t+1)$-th meta-training data set $\dataset_{t+1}$. We adopt the general principle of maximizing the amount of ``knowledge'' that can be extracted from the data set associated with selected channel $c_{t+1}$, when added to the $t$ available data sets $\dataset_{1:t}$. This is done via the following three steps: \emph{(i)} searching in the space of model parameter vectors for a vector $\phi_{t+1}$ that is most ``surprising'' given the available meta-training data $\dataset_{1:t}$; \emph{(ii)} translating the selected model parameter vector $\phi_{t+1}$ into a channel $c_{t+1}$; and \emph{(iii)} generating data set $\dataset_{t+1}$ by using the simulator with input $c_{t+1}$.

\begin{figure*}%[!ht]
    \centering
    \includegraphics[page=2,trim = 0cm 0cm 0cm 0cm, clip, width=0.95\textwidth]{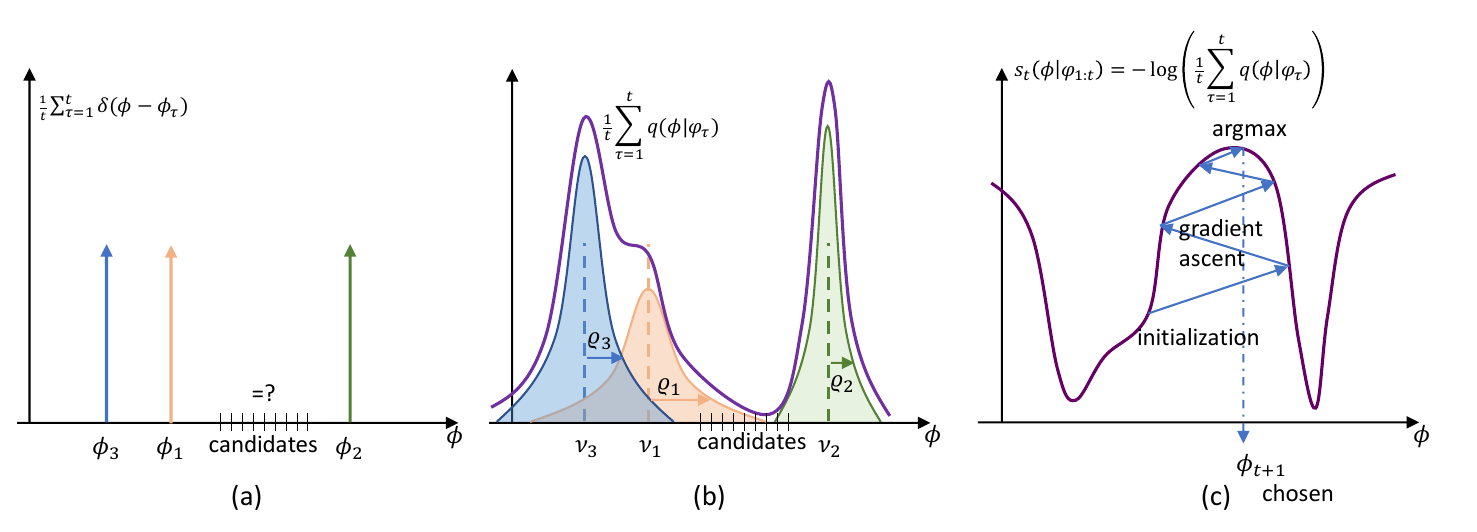}
    \caption{Illustration of how model parameter vectors are scored to enable active meta-learning provided $t=3$ meta-training sets. (a) Frequentist meta-learning relies on point estimates, and is hence unable to score as-of-yet unexplored model parameters; (b) Bayesian meta-learning can associate a score to each model parameter vector $\phi$ based on the variational distributions $\big\{q(\phi|\varphi_\tau)\big\}$ evaluated in the previously observed frames $\tau=1,\dots,t$; (c) The scoring function can be maximized to obtain the next model parameter vector $\phi_{t+1}$ as the most ``surprising'' one.}
    \label{fig: Illustration of Bayesian scoring over model parameter space}
\end{figure*}

As illustrated in Fig.~\ref{fig: Illustration of Bayesian scoring over model parameter space}, in step \emph{(i)}, we adopt the scoring function introduced in \cite{kaddour2020probabilistic}, i.e.,
\begin{equation}
    s_t(\phi|\varphi_{1:t}) := -\log \Bigg( \tfrac{1}{t} \sum_{\tau=1}^t q(\phi|\varphi_\tau) \Bigg) \label{eq: s_t}
\end{equation}
in order to select the next model parameter vector as
\begin{equation}
    \phi_{t+1}=\argmax_{\phi} s_t(\phi|\varphi_{1:t}). \label{eq: phi_t+1 = argmax s}
\end{equation}
The criterion \eqref{eq: s_t} measures how incompatible model parameter vector $\phi$ is with the available data $\dataset_{1:t}$. In fact, by the derivations in the previous section: the mixture of variational distributions $\tfrac{1}{t} \sum_{\tau=1}^t q(\phi|\varphi_\tau)$ quantifies how likely a vector $\phi$ is on the basis of the data $\dataset_{1:t}$ (Fig. \ref{fig: Illustration of Bayesian scoring over model parameter space}b); and the negative logarithm in \eqref{eq: s_t} evaluates the information-theoretic ``surprise'' associated with that mixture. Problem \eqref{eq: phi_t+1 = argmax s} can be addressed either by grid search for low-dimensional model parameter space, or by using gradient ascent due to the differentiability nature of the scoring function \eqref{eq: s_t}, as illustrated in Fig. \ref{fig: Illustration of Bayesian scoring over model parameter space}c.

In step \emph{(ii)}, we need to convert the selected model parameter vector $\phi_{t+1}$, i.e., the outcome of \eqref{eq: phi_t+1 = argmax s}, into channel state $c_{t+1}$. We choose the channel state $c_{t+1}$ that minimizes the cross entropy loss when evaluated at $\phi_{t+1}$, i.e.,
\begin{equation}
    c_{t+1} \in \argmin_c \big\{ \mathcal{L}_\text{p}(\phi_{t+1}|c) = \E_{p(x,y|c)}[-\log p(x|y,\phi_{t+1})] \big\}, \label{eq: c_{t+1} = argmin L_p(phi|c) } 
\end{equation}
where we set $p(x,y|c)=p(x)p(y|x,c)$, with $p(x)$ being some fixed distribution and $p(y|x,c)$ being the distribution of the output of the simulator. In \eqref{eq: c_{t+1} = argmin L_p(phi|c) }, we have emphasized that there may be more than one solution to the problem.
The rational behind problem \eqref{eq: c_{t+1} = argmin L_p(phi|c) } is that data generated from the distribution $p(x,y|c_{t+1})$ can be interpreted as being the most compatible with the demodulator $p(x|y,\phi_{t+1})$, where compatibility is measured by the average of the cross entropy $\E_{p(y|c_{t+1})} \big[\mathrm{H}\big(p(x|y,c_{t+1}),p(x|y,\phi_{t+1})\big)\big]$. 

We emphasize that the proposed approach is different from the methodology introduced by \cite{kaddour2020probabilistic}, which uses another variational distribution in problem \eqref{eq: Bayesian meta-learning as optim problem}. In our experiments, we found the method in \cite{kaddour2020probabilistic} to be ineffective and complex for the problem under study here. The main issue appears to be overfitting for the additional variational distribution, which is overcome by leveraging the availability of the channel simulator implementing the model $p(y|x,c)$.

In some models, problem \eqref{eq: c_{t+1} = argmin L_p(phi|c) } can be solved analytically. For more complex models, SGD-based approaches can be used, either by differentiating an estimate of the loss in a manner similar to the discussion in Sec.~\ref{sec: Bayesian VI ML} $\big($i.e., Algorithm~\ref{alg: reparm trick} with $\mathcal{G}(\phi_{t+1})=\mathcal{L}_\text{p}(\phi_{t+1}|c)\big)$, or by directly estimating its gradient \cite{mohamed2020monte}.

Finally, in step \emph{(iii)}, meta-training data set $\dataset_{t+1}=\{(y_{t+1}[i],x_{t+1}[i])\}_{i=1}^{N_{t+1}}$ is generated using the simulator in an i.i.d. fashion following the distribution
\begin{equation}
    \prod_{i=1}^{N_{t+1}} p(x_{t+1}[i]) p\big(y_{t+1}[i] \big| x_{t+1}[i], c_{t+1}\big). \label{eq: generate D_{t+1}}
\end{equation}
As a final note, we adopt the proposal in \cite{kaddour2020probabilistic} of implementing active selection only after $t_\text{init}>1$ channel states that are generated at random, as a means to avoid being overconfident at early stages. The overall proposed Bayesian active meta-learning scheme is summarized in Algorithm \ref{alg: bayesian active meta training}.

\begin{algorithm}
    \caption{\\ \texttt{Bayesian Active Meta-Training}}
    \label{alg: bayesian active meta training}
    \KwInputs{      $p(y|x,c)$ = channel model\\
                    $p(x)$ = generative symbols distribution\\}
    \KwParameters{  $t_\text{init}$ = number of prior-based first frames
                    }
    \KwOutput{      $\xi$ = shared hyperparameter vector
                    }
    \BlankLine
    \algcomment{Generate initial experience}\;
    \For{$t= 1,2,\dots,t_\text{init}$}{
        Draw using the prior $c_t \sim p(c_t)$ \;
        Acquire data $\dataset_t \sim \prod_{i=1}^{N_t} p(x_t[i]) p(y_t[i]|x_t[i],c_t)$\;
    }
    \While{data acquisition not done}{
        $\xi \gets \mathtt{BayesianMetaTraining}(\dataset_{1:t})$ using Algorithm~\ref{alg: bayesian maml meta-training}\;
        \algcomment{frame-specific update with updated $\xi$}\;
        \For{$\tau=1,2,\dots,t$}{
            $\varphi_\tau \gets \varphi^\text{GD}(\dataset_\tau^\text{tr}|\xi)$ using \eqref{eq: F_hat_D_tau(varphi,xi) def}\;
        }
        \algcomment{step \emph{(i)}, choose surprising model parameter}\;
        $\phi_{t+1}=\argmax_{\phi} s_t(\phi|\varphi_{1:t})$ using \eqref{eq: s_t} \;
        \algcomment{step \emph{(ii)}, choose next channel}\;
        $c_{t+1} \in \argmin_c \mathcal{L}_\text{p}(\phi_{t+1}|c)$ as in \eqref{eq: c_{t+1} = argmin L_p(phi|c) } \;
        \algcomment{step \emph{(iii)}, generate data set}\;
        $\dataset_{t+1} \sim \prod_{i=1}^{N_{t+1}} p(x_{t+1}[i]) p(y_{t+1}[i]|x_{t+1}[i],c_{t+1})$ \label{algline: bayesiandata generation} \;
        $t \gets t+1$ \;
    }
    \Return $\xi$
\end{algorithm}

%%%%%%%%%%%%%%%%%%%%%%%%%%%%%%%%%%%%%%%%%%%%%%%%%%%%%%%%%%%%%%%%%%%%%%%%%%%%%%%%%%%%%%%%%%%%%%%%%%%%%%%%%%
%%%%%%%%%%%%%%%%%%%%%%%%%%%%%%%%%%%%%%          Section             %%%%%%%%%%%%%%%%%%%%%%%%%%%%%%%%%%%%%%
%%%%%%%%%%%%%%%%%%%%%%%%%%%%%%%%%%%%%%%%%%%%%%%%%%%%%%%%%%%%%%%%%%%%%%%%%%%%%%%%%%%%%%%%%%%%%%%%%%%%%%%%%%
\section{Experiments}\label{sec: Experiments}
In this section, we present experimental results to evaluate the performance of Bayesian meta-learning for demodulation/equalization.

%%%%%%%%%%%%%%%%%%%%%%%%%%%%%%%%%%%%%%%%%%%%%%%%%%%%%%%%%%%%%%%%%%%%%%%%%%%%%%%%%%%%%%
%%%%%%%%%%%%%%%%%%%%%%%%%%%%        Subsection            %%%%%%%%%%%%%%%%%%%%%%%%%%%%
%%%%%%%%%%%%%%%%%%%%%%%%%%%%%%%%%%%%%%%%%%%%%%%%%%%%%%%%%%%%%%%%%%%%%%%%%%%%%%%%%%%%%%
\subsection{Performance Metrics} \label{sec:perf_metric}

Apart from the standard measures of symbol error rate (SER) and mean squared error (MSE), we will also evaluate metrics quantifying the performance in terms of the reliability of the confidence measures provided by the predictor. While such measures can be defined for both classification and regression problems, we will focus here on uncertainty quantification for demodulation via calibration metrics (see \cite{melluish2001comparing} for discussion on regression). 

As discussed in the previous sections, for a new frame, we need to make a prediction for the payload symbols $\{y_*^\text{te}[i]\}_{i=1}^{N_*^\text{te}}$ via the demodulator $p(x_*^\text{te}[i]|y_*^\text{te}[i],\phi_*)$ for frequentist meta-learning \eqref{eq:freqpred}, or $p(x_{*}^\text{te}[i]|y_{*}^\text{te}[i],\varphi_*) $ for Bayesian meta-learning \eqref{eq:bayesian_ensemble_prediction}. The \emph{confidence level} assigned by the model to the hard predicted symbol
\begin{align}
    \hat{x}_*^\text{te}[i] = \argmax_{x \in \mathcal{X}} p(x|y_*^\text{te}[i],\theta)
\end{align}
given the received symbol $y_*^\text{te}[i]$, can be defined as the corresponding probability \cite{Guo2017Calibration}
\begin{align}
    \hat{p}[i] = \max_{x \in \mathcal{X}} p(x|y_*^\text{te}[i],\theta) =  p(\hat{x}_*^\text{te}[i]|y_*^\text{te}[i],\theta),
\end{align}
where we have $\theta = \phi_*$ for frequentist meta-learning and $\theta=\varphi_*$ for Bayesian meta-learning. \emph{Perfect calibration} \cite{Guo2017Calibration} can be defined as the condition where symbols that are assigned a confidence level $\hat{p}[i]$ are also characterized by a probability of correct detection equal to $p$.

Two standard means of quantifying the extent to which the perfect calibration is satisfied are reliability diagrams \cite{degroot1983comparison} and expected calibration error (ECE) \cite{Guo2017Calibration}. To introduce them, the probability interval $[0,1]$ is first divided into $M$ equal length intervals, with the $m$-th interval $(\tfrac{m-1}{M}, \tfrac{m}{M}]$ referred to as the $m$-th bin henceforth. Let us denote as $\mathcal{B}_m$ the subset of the payload data symbol indices whose associated confidence level $\hat{p}[i]$ lie within the $m$-th bin, i.e.,
\begin{equation}
    \mathcal{B}_m = \big\{ i \big| \hat{p}[i] \in \big(\tfrac{m-1}{M}, \tfrac{m}{M}\big],\textrm{ with } i=1,2,\dotsc,N_*^\text{te}\big\}.
\end{equation}
Note this is a partition of the data set $\dataset_*^\text{te}$ since we have  $\bigcup_{m=1}^M \mathcal{B}_m~=~ \{i~=~1,2,\dotsc,N_*^\text{te}\}$ and $\mathcal{B}_m \cap \mathcal{B}_{m'} =\emptyset$ for any $m' \neq m$.

The within-bin empirical average accuracy of the predictor for the $m$-th bin is defined as
\begin{align}
    \mathrm{acc}(\mathcal{B}_m) = \frac{1}{|\mathcal{B}_m|} \sum_{i \in \mathcal{B}_m} \mathbf{1}(\hat{x}_*^\text{te}[i] = x_*^\text{te}[i]),
    \label{eq: acc(B_m)}
\end{align}
with $\mathbf{1}(\cdot)$ being indicator function and $|\mathcal{B}_m|$ denoting the number of total samples in $\mathcal{B}_m$. The within-bin empirical average confidence of the predictor for the $m$-th bin is 
\begin{align}
    \mathrm{conf}(\mathcal{B}_m) = \frac{1}{|\mathcal{B}_m|} \sum_{i \in \mathcal{B}_m}\hat{p}[i].
    \label{eq: conf(B_m)}
\end{align}
A perfectly calibrated demodulator $p(x|y,\theta)$ would have $\mathrm{acc}(\mathcal{B}_m) = \mathrm{conf}(\mathcal{B}_m)$ for all $m \in \{1,\ldots,M\}$ in the limit of a sufficiently large payload data set, i.e., $N_*^\text{te} \rightarrow \infty$. 

Reliability diagrams plot the accuracy $\mathrm{acc}(\mathcal{B}_m)$ and the confidence $\mathrm{conf}(\mathcal{B}_m)$ over the binned probability interval $[0,1]$. Ideal calibration would yield $\mathrm{acc}(\mathcal{B}_m) = \mathrm{conf}(\mathcal{B}_m)$ in a reliability plot. If in the $m$-th bin, the empirical accuracy and empirical confidence are different, the predictor is considered to be over-confident when $\mathrm{conf}(\mathcal{B}_m) > \mathrm{acc}(\mathcal{B}_m)$, and  under-confident when $\mathrm{conf}(\mathcal{B}_m) < \mathrm{acc}(\mathcal{B}_m)$.

The ECE quantifies the overall amount of miscalibration by computing the weighted average of the differences between within-bin accuracy and within-bin confidence levels across all $M$ bins, i.e.,
\begin{align}
    \text{ECE} = \frac{1}{N_*^\text{te}}\sum_{m=1}^M \big| \mathcal{B}_m \big| \Big| \mathrm{acc}(\mathcal{B}_m) - \mathrm{conf}(\mathcal{B}_m) \Big|.
\end{align}

%%%%%%%%%%%%%%%%%%%%%%%%%%%%%%%%%%%%%%%%%%%%%%%%%%%%%%%%%%%%%%%%%%%%%%%%%%%%%%%%%%%%%%
%%%%%%%%%%%%%%%%%%%%%%%%%%%%        Subsection            %%%%%%%%%%%%%%%%%%%%%%%%%%%%
%%%%%%%%%%%%%%%%%%%%%%%%%%%%%%%%%%%%%%%%%%%%%%%%%%%%%%%%%%%%%%%%%%%%%%%%%%%%%%%%%%%%%%
\subsection{Frequentist and Bayesian Meta-Learning for Demodulation}\label{sec: Experiments Demodulation}

For the first set of experiments, we focus on a demodulation problem at the symbol level in the presence of transmitter I/Q imbalance \cite{zhang2020deepwiphy, helmy2017robustness}, as considered also in \cite{park2021fewpilots}. The main reason for this choice is that channel decoding typically requires a hard decision on the transmitted codeword, whose accuracy can be validated via a cyclic redundancy check. In contrast, demodulation is usually a preliminary step at the receiver side, and downstream blocks, such as channel decoding, expect soft inputs that are well calibrated. For each frame $\tau$, the transmitted symbols $x_\tau[i]$ are drawn uniformly at random from the 16-QAM constellation $\mathcal{X}=1/\sqrt{10} (\{\pm1,\pm3\}+\jmath\{\pm1,\pm3\})$. The received symbol $y_\tau[i] \in \mathcal{Y}=\mathbb{C}$ is given as 
\begin{equation}
    y_\tau[i] = h_\tau f_{\text{IQ},\tau}({x}_\tau[i]) + z_\tau[i], \label{eq: demodulation channel model}
\end{equation}
for a unit energy fading channel coefficient $h_\tau$, where the additive noise is $ z_\tau[i] \sim \mathcal{CN}(0,\SNR^{-1}) $ for some signal-to-noise ratio (SNR) level $\SNR$, and the I/Q imbalance function \cite{tandur2007joint} $f_{\text{IQ},\tau}: \mathcal{X} \rightarrow \bar{\mathcal{X}}_\tau$ is
\begin{eqnarray}
    f_{\text{IQ},\tau}({x}_\tau[i]) 
    \negspaceF&=&\negspaceF \bar{x}_{\text{I},\tau}[i] + \jmath \bar{x}_{\text{Q},\tau}[i] \label{eq:iq_imbalance} \\
    \begin{bmatrix} 
        \bar{x}_{\text{I},\tau}[i] \\ \bar{x}_{\text{Q},\tau}[i]
    \end{bmatrix}
    \negspaceF&=&\negspaceF
    \begin{bmatrix} 
        1+\epsilon_\tau & 0 \\ 0 & 1-\epsilon_\tau
    \end{bmatrix}
    \begin{bmatrix} 
        \cos \delta_\tau & -\sin \delta_\tau \\ -\sin \delta_\tau & \cos \delta_\tau
    \end{bmatrix}
    \begin{bmatrix} 
        x_{\text{I},\tau}[i] \\ x_{\text{Q},\tau}[i]
    \end{bmatrix}, \nonumber
\end{eqnarray}
which depends on the imbalance parameters $\epsilon_\tau$ and $\delta_\tau$.
In \eqref{eq: demodulation channel model}, $x_{\text{I},\tau}[i] $ and $x_{\text{Q},\tau}[i]$ refer to the real and imaginary parts of the modulated symbol $x_\tau[i]$; and  $\bar{x}_{\text{I},\tau}[i]$  and $\bar{x}_{\text{Q},\tau}[i]$ stand for the real and imaginary parts of the transmitted symbol $f_{\text{IQ},\tau}(x_\tau[i])$. Note that the constellation $\bar{\mathcal{X}}_\tau$ of the transmitted symbols $\bar{x}_\tau[i]$ is also composed of $16$ points via \eqref{eq:iq_imbalance}.

By \eqref{eq: demodulation channel model} and \eqref{eq:iq_imbalance}, the channel state $c_\tau$ consists of the tuple: (a) amplitude imbalance factor $\epsilon_\tau\in [0,0.15]$; (b) phase imbalance factor $\delta_\tau\in [0,15^\circ]$; and (c) channel realization $h_\tau\in\Cset$. All of the variables are drawn i.i.d. across different frames and are fixed during each frame. We consider the channel state distribution for frame $\tau$ as
\begin{equation}
    p(c_\tau)  = \Betadist\Big(\tfrac{\epsilon_\tau}{0.15} \Big| 5,2\Big) \Betadist\Big(\tfrac{\delta_\tau}{0.15^{\circ}} \Big| 5,2\Big) \ComplexNormdist(h_\tau | 0,1).
\end{equation}  

We set our base learner to be a multi-layer fully-connected neural network \eqref{eq: p(x|y,phi) model} with $L=5$ layers. The real and imaginary parts of the input $y[i]\in\mathbb{C}$ are treated as a vector in $\mathbb{R}^2$, which is fed to  layers with $10$, $30$, and $30$ neurons, all with ReLU activations, while the last linear layer implements a softmax function that produces probabilities for the 16QAM constellation points.

\begin{figure}[ht!]
    \centering
    \includegraphics[trim=0cm 0.1cm 0cm 0.8cm, clip, width=9cm]{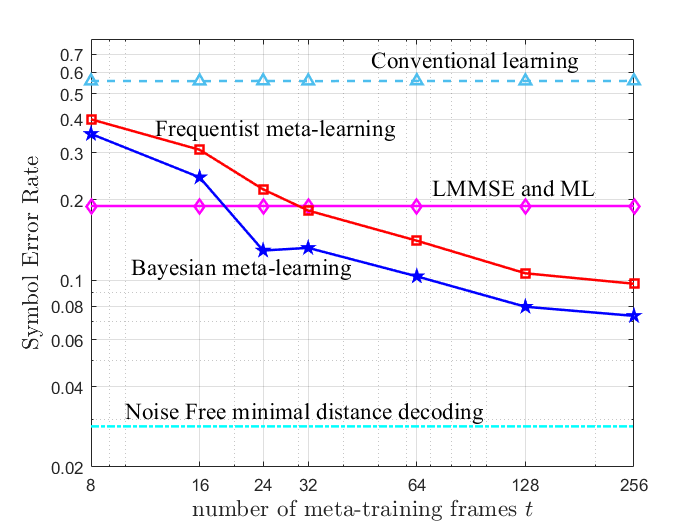}
    \caption{Symbol error rate (SER) as a function of the number $t$ of meta-training frames with 16-QAM, Rayleigh fading, and I/Q imbalance for $N_\tau^\text{tr}=4$, $N_*^\text{tr}=8$. The symbol error rate is averaged over by $N_*^\text{te}=4000$ data symbols and $50$ meta-test frames with ensemble of size $100$.}
    \label{fig: SER_vs_num_mtr_frames}
    %\vspace{-0.5cm}
\end{figure}

To address the ability of meta-learning to adapt the demodulator using only few pilots, we set the number of pilots as $N_\tau^\text{tr}=4$ during meta-training and $N_*^\text{tr}=8$ for meta-testing \cite{park2021fewpilots}. Fig.~\ref{fig: SER_vs_num_mtr_frames} shows the SER as a function of the number of total meta-training frames $t$. Since only half of the constellation points are available as pilots during meta-test ($N_*^\text{tr}=8$ different symbols out of $16$), conventional learning cannot obtain a SER lower than of 0.5. In fact, conventional learning performs worse than a standard model-based receiver applying linear minimal mean square error (LMMSE), followed by maximum likelihood (ML) demodulation, while disregarding the presence of I/Q imbalance function $f_{\text{IQ}}$. Both meta-learning schemes are clearly superior to conventional learning and to the mentioned model-based solution, showing that useful knowledge has been transferred from previous frames to a new frame. Furthermore, Bayesian meta-learning obtains a slightly lower SER as compared to  frequentist meta-learning. This advantage stems from the capacity of ensemble predictors to implement more complex decision boundaries \cite{zecchin2022robust}.

\begin{figure}%[!ht]
    \centering
    \includegraphics[trim=0cm 0.1cm 0cm 0.6cm, clip, width=9cm]{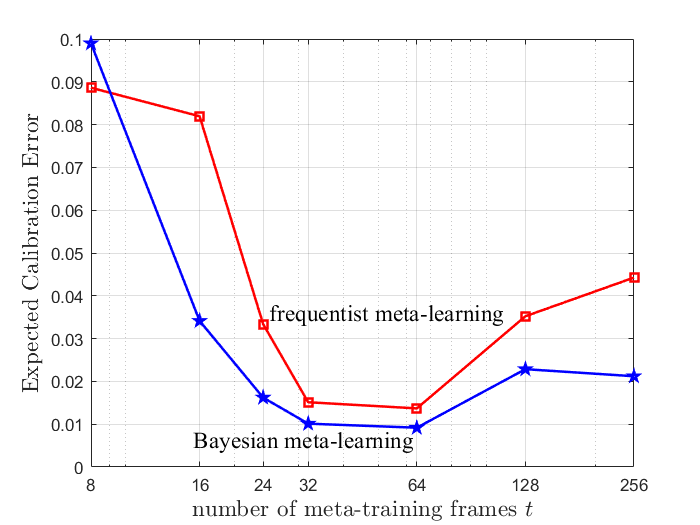}
    \caption{Expected calibration error (ECE) over meta-test data $\dataset_*^\text{te}$ as a function of the number $t$ of meta-training frames, for the same setting as in Fig.~\ref{fig: SER_vs_num_mtr_frames}.}
    \label{fig: ECE_vs_num_mtr_frames}
\end{figure}

To gain insights into the reliability of the uncertainty quantification provided by the demodulator, we use the metrics defined in Sec.~\ref{sec:perf_metric} by setting the total number of bins to $M=10$. We plot the ECE as a function of the number of total meta-training frames $t$ in Fig.~\ref{fig: ECE_vs_num_mtr_frames}.  Bayesian meta-learning is seen to achieve a  lower ECE than frequentist meta-learning, indicating that Bayesian meta-learning provides more reliable estimates of uncertainty. Furthermore, the increase in ECE as the number $t$ of available meta-training frames increases may be interpreted as a consequence of meta-overfitting \cite{yin2019meta}. This suggests that meta-learning may be considered as complete after a number of frames that depends on the complexity of the propagation environment. In practice, this can be assessed by evaluating the performance of the demodulator on pilots (see the online strategy in \cite{park2021fewpilots} for further discussion on this point).

\begin{figure}%[!ht]
    \centering
    \includegraphics[width=4.3cm]{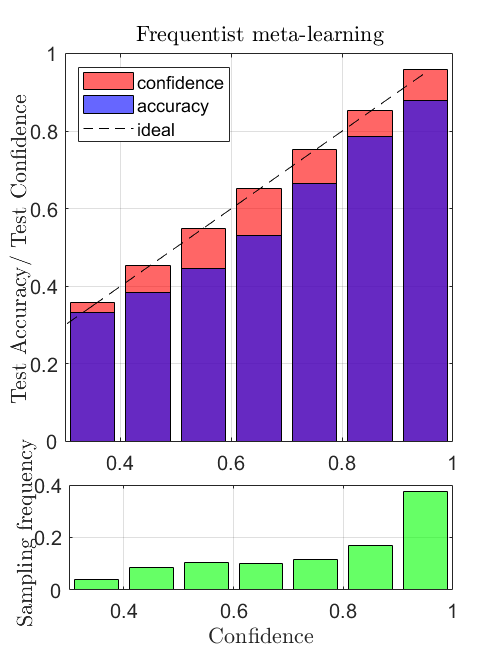}
    \includegraphics[width=4.3cm]{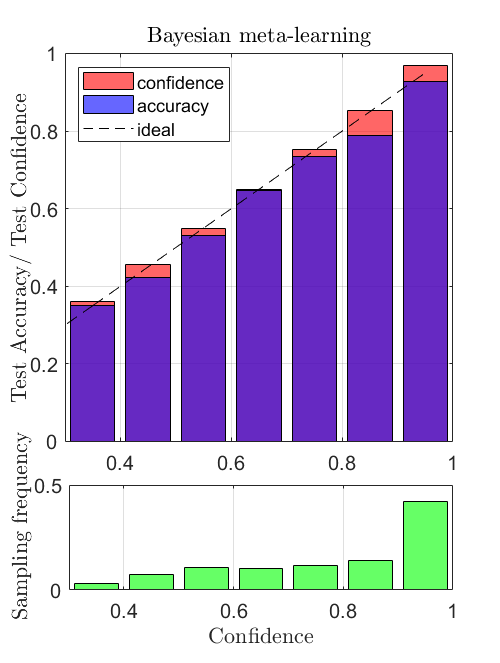}
    \vspace{-0.15cm}
    \caption{Reliability diagrams (top) for frequentist meta-learning (left) and Bayesian  meta-learning (right) with $\SNR=18$ dB, using $t=16$ meta-training frames and predictions averaged over $50$ meta-test frames. Frequentist meta-learning tends to be over-confident, whereas the Bayesian soft predictions are better matched to the true accuracy. The bottom figure shows the histogram of $|\mathcal{B}_m|/N$ of prediction over $M=10$ bins. Full details in Appendix~\ref{appendix: Experiments Details}}
    \label{fig: reliability_diagrams}
    %\vspace{-0.55cm}
\end{figure}

To further elaborate on the quality of uncertainty quantification, Fig.~\ref{fig: reliability_diagrams} depicts reliability diagrams for frequentist and Bayesian meta-learning. The within-bin accuracy levels $\mathrm{acc}(\mathcal{B}_m)$ in \eqref{eq: acc(B_m)} and the within-bin empirical confidence $\mathrm{conf}(\mathcal{B}_m)$ in \eqref{eq: conf(B_m)} are depicted as dark (blue) and light (red) bars, respectively. Frequentist meta-learning is observed to produce generally over-confident predictions, while Bayesian meta-learning provides better calibrated predictions with well-matching confidence and accuracy levels. 

%%%%%%%%%%%%%%%%%%%%%%%%%%%%%%%%%%%%%%%%%%%%%%%%%%%%%%%%%%%%%%%%%%%%%%%%%%%%%%%%%%%%%%
%%%%%%%%%%%%%%%%%%%%%%%%%%%%        Subsection            %%%%%%%%%%%%%%%%%%%%%%%%%%%%
%%%%%%%%%%%%%%%%%%%%%%%%%%%%%%%%%%%%%%%%%%%%%%%%%%%%%%%%%%%%%%%%%%%%%%%%%%%%%%%%%%%%%%
\subsection{Bayesian Active Meta-Learning for Equalization}\label{sec: Experiments Equalization}
In this sub section, we illustrate the operation of active meta-learning by investigating a single-input multiple-output (SIMO) Rayleigh block fading real channel model. At frame $\tau$, the modulator uses a 4-PAM to produce symbols $x_\tau[i]$, $i=1,2,\dots,N_\tau,$ taken uniformly from the set $\mathcal{X} \in{1/\sqrt{5}}\{-3,-1,+1,+3\}$. Given channel state $c_\tau =[c_{\tau}^0,c_{\tau}^1]^\top \in \mathbb{R}^2$, the $i$-th channel output symbol $y_\tau[i]\in\mathbb{R}^2$ for $i=1,2,\dots,N_\tau$ is defined as the two-dimensional real vector
\begin{equation}
    y_\tau[i] =  c_\tau x_\tau[i] + z_\tau[i], \label{eq: channel model of equalization}
\end{equation}
where both the additive noise $z_\tau[i] \sim \mathcal{N}\big(0, \tfrac{1}{2\SNR} I_2\big)$ and the normalized real block fading coefficients $c_\tau \sim p(c)=\Normdist(c|0,I_2)$ are i.i.d. We adopt the linear equalizer
\begin{equation}
    \hat{x}_\tau[i] = \phi_\tau^\top \cdot y_\tau[i] \label{eq: linear equalizer x = phi^T y}
\end{equation}
with linear equalizer weight vector $\phi_\tau=[\phi_\tau^0,\phi_\tau^1]^\top\in\Rset^2$. To obtain a soft equalization, we account for a precision level \changemade{$\beta$} via the conditional distribution 
\begin{equation}
    p(x_\tau[i]|y_\tau[i],\phi_\tau) = \mathcal{N}(\phi_\tau^\top \cdot y_\tau[i], \beta^{-1}). \label{eq: soft equalization}
\end{equation}

The next model parameter $\phi_{t+1}$ is chosen to maximize the scoring function as in \eqref{eq: phi_t+1 = argmax s} by restricting the optimization  to the domain $||\phi|| \leq 1$. This restricted optimization domain is selected in order to match the circular symmetry of the problem. Furthermore, the corresponding next channel state $c_{t+1}$ is selected by tackling problem \eqref{eq: c_{t+1} = argmin L_p(phi|c) }, which amounts to the minimization
\begin{subequations}
    \begin{eqnarray}
        c_{t+1}(\phi) 
        &\in& \argmin_c \E_{p(x)p(y|x,c)}[- \log p(x|y,\phi)] \\
        &=& \argmin_c \changemade{\mathrm{E}_{p(x)p(z)}[\tfrac{\beta}{2} (x-\phi^\top \cdot(c x + z ))^2]} \nonumber \\
        &=& \argmin_c \changemade{\mathrm{E}_{p(x)p(z)}[\tfrac{\beta}{2} \big( (1-\phi^\top \cdot c) x  - \phi^\top z \big)^2]}
        \nonumber \\
        &=& \big\{ c \big| \phi^\top \cdot c = 1 \big\}. \label{eq: min norm c_t+1(phi)}
    \end{eqnarray}
\end{subequations}
In the set of solutions of problem \eqref{eq: min norm c_t+1(phi)}, we select the minimum-norm solution $c_{t+1}=\phi_{t+1} / \|\phi_{t+1}\|^2$. This way, the selected channel focuses on the more challenging low-SNR regime. Details of this experiment are provided in Appendix~\ref{appendix: Experiments Details}.

Fig.~\ref{Fig: scoring function of active} illustrates the scoring function \eqref{eq: s_t} used to select the next model parameter $\phi_{t+1}$ as a heat map in the space of model parameter $\phi$. Specifically, the figure shows the scoring functions after observing $t=4$ and $t=5$ meta-training frames. The optimized next model parameter vector $\phi_{t+1}$ \eqref{eq: phi_t+1 = argmax s} is shown as a star, while the previously selected model parameter vectors $\phi_{1:t}$ are shown as squares. Fig.~\ref{Fig: scoring function of active} illustrates how active meta-learning efficiently explores the model parameter space. It does so by avoiding the inclusion of channel states that are similar to those already considered (i.e., the squares in the figure). This way, the model parameter space can be covered with fewer meta-training frames $t$, leading to a larger frame efficiency of active meta-learning. 

\begin{figure}%[!ht]
    \centering
    \includegraphics[trim = 0.5cm 2.0cm .5cm 0.5cm, clip,width=9cm]{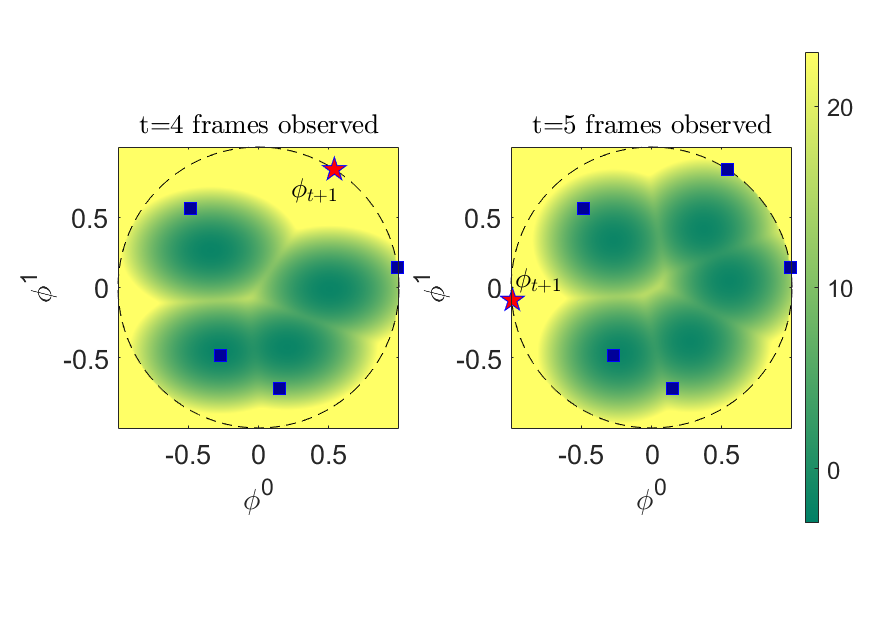}
    \vspace{-0.5cm}
    \caption{Scoring function \eqref{eq: s_t} used by Bayesian active meta-learning to select the next model parameter vector $\phi_{t+1}$ at the fourth and fifth iterations. The scoring function is shown as a heat map over the two dimensional space of the model parameter vector $\phi$ for the example detailed in Sec.~\ref{sec: Experiments Equalization}. }
    \label{Fig: scoring function of active}
\end{figure}

\begin{figure}%[!ht]
    \centering
    \includegraphics[page=1,trim=0cm 0cm 0cm 0cm, clip,width= 9cm]{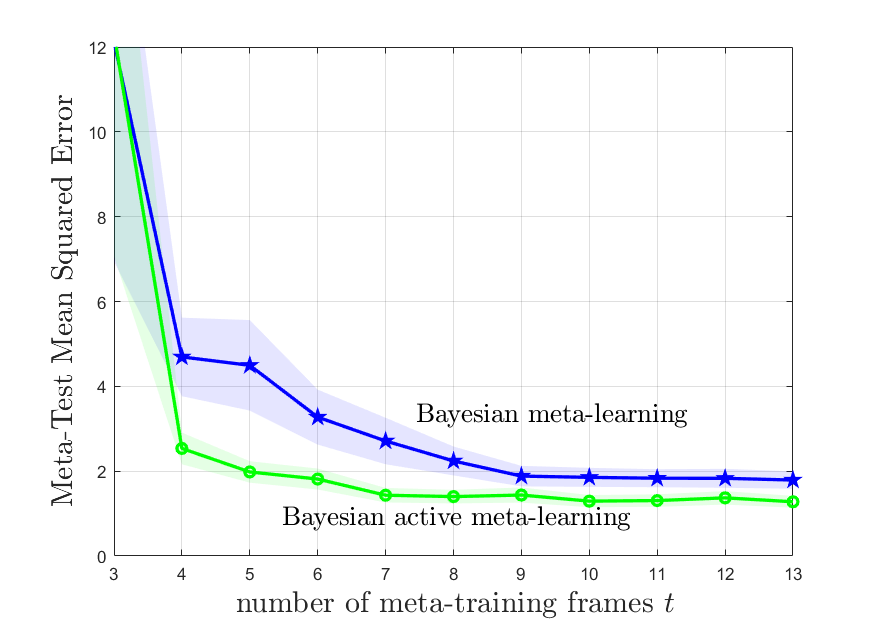}
    \caption{Meta-test mean squared error (MSE) loss as function of the number of frames $t$. Bayesian active meta-training is able to achieve lower meta-test loss levels by using fewer meta-training tasks $t$. Solid lines are the mean test loss over 100 channel states. The confidence levels account for one standard deviation.}
    \label{fig: meta-test loss vs. num meta-training frames}
    \vspace{-0.5cm}
\end{figure}

Finally, to numerically validate the advantage of active meta-learning, we plot the meta-test MSE loss in Fig.~\ref{fig: meta-test loss vs. num meta-training frames} for both passive and active Bayesian meta-learning versus the number of frames $t$. For passive meta-learning, we have generated random channel realizations by drawing from the distribution $p(c)=\mathcal{N}(c|0,I_2)$. We have repeated the experiment $100$ times, and show the confidence interval  of one standard deviation for the meta-test loss. The results in the figure confirm  that active meta-learning requires far fewer meta-training frames. Furthermore, the increased randomness of passive meta-learning is due to the random selection of channel states at each iteration.

%%%%%%%%%%%%%%%%%%%%%%%%%%%%%%%%%%%%%%%%%%%%%%%%%%%%%%%%%%%%%%%%%%%%%%%%%%%%%%%%%%%%%%%%%%%%%%%%%%%%%%%%%%
%%%%%%%%%%%%%%%%%%%%%%%%%%%%%%%%%%%%%%          Section             %%%%%%%%%%%%%%%%%%%%%%%%%%%%%%%%%%%%%%
%%%%%%%%%%%%%%%%%%%%%%%%%%%%%%%%%%%%%%%%%%%%%%%%%%%%%%%%%%%%%%%%%%%%%%%%%%%%%%%%%%%%%%%%%%%%%%%%%%%%%%%%%%
%\vspace{-0.05cm}
\section{Conclusions}
\label{sec: conclusions}
%\vspace{-0.05cm}
In this paper, we have introduced tools for reliable and efficient AI in communication systems via Bayesian meta-learning. Bayesian learning has the advantage of producing well-calibrated decisions whose confidence levels are a close match for the corresponding test accuracy. This property facilitates  monitoring of the quality of the outputs of an AI module. Meta-learning optimizes models that can quickly adapt based on few pilots, producing sample-efficient AI solutions. This paper has focused on the application of Bayesian meta-learning to the basic problems of demodulation/equalization from few pilots.  We have demonstrated via experiments that the demodulator/equalizer obtained via Bayesian meta-learning not only achieves a higher accuracy, but it also enjoys better calibration performance than its standard frequentist counterpart. Furthermore, thanks to meta-learning, such performance levels can be obtained based on a limited number of pilots per frame. 

To reduce the number of past frames required by meta-learning, we have also introduced Bayesian active meta-learning, which leverages the uncertainty estimates produced by Bayesian learning to actively explore the space of channel conditions. We have shown via numerical results that active meta-learning can indeed significantly speed up meta-training in terms of number of frames.

Future work may consider a fully Bayesian meta-learning implementation that also accounts for uncertainty at the level of hyperparameters (see, e.g., \cite{jose2022information} and references therein). This may be particularly useful in the regime of low number of frames. Another direction for research would be to investigate different scoring functions for active meta-learning (see, e.g., \cite{nikoloska2021bamldbo}). \changemade{A study on the impact of well-calibrated decisions obtained via Bayesian learning on downstream blocks at the receiver, such as channel decoding, is also of interest.} Finally, the proposed tools may find applications to other problems in communications, such as power control \cite{nikoloska2021fast} and channel coding \cite{jiang2019mind, li2021channel}.

%%%%%%%%%%%%%%%%%%%%%%%%%%%%%%%%%%%%%%%%%%%%%%%%%%%%%%%%%%%%%%%%%%%%%%%%%%%%%%%%%%%%%%%%%%%%%%%%%%%%%%%%%%
%%%%%%%%%%%%%%%%%%%%%%%%%%%%%%%%%%%%%%          Appendix             %%%%%%%%%%%%%%%%%%%%%%%%%%%%%%%%%%%%%%
%%%%%%%%%%%%%%%%%%%%%%%%%%%%%%%%%%%%%%%%%%%%%%%%%%%%%%%%%%%%%%%%%%%%%%%%%%%%%%%%%%%%%%%%%%%%%%%%%%%%%%%%%%
\appendices

\begin{table*}[tbp]
    \centering
    \caption{Parameters for the demodulation and equalization meta-learning.}
    \label{tab: parameters used in this paper}
    \begin{tabular}{l  c c c } \toprule
        Description & Symbol & Demodulation (Sec.~\ref{sec: Experiments Demodulation}) & Equalization (Sec.~\ref{sec: Experiments Equalization}) \\ [0.5ex] \midrule\midrule
        Signal to noise ratio [dB]                                      & $\SNR$                        & $18$                  & $6$                       \\
        Modulation                                                      & -                             & 16-QAM                & 4-PAM                     \\
        Neural network input dimension                                  & integer                       & $\dim(\mathbb{C})=2$  & $\dim(\mathbb{R}^2)=2$    \\
        Neural network layers size [\{hidden\},output]                  & neurons per layer             & $[10,30,30,16]$       & $[,1]$                    \\
        Activation of hidden layers                                     & -                             & ReLU                  & NA                        \\
        Activation of last layer                                        & -                             & softmax               & no activation             \\
        Meta-training frames mini-batch size                            & $B$                           & $16$                  & full batch ($t$)          \\
        Frame-specific learning rate                                    & $\eta$                        & $10^{-1}$             & $2\cdot10^{-3}$           \\
        Meta-learning rate                                              & $\kappa$                      & $10^{-3}$             & $5\cdot10^{-2}$           \\
        Number of pilots for frame-specific updates while meta-training & $N_\tau^\text{tr}$            & $4$                   & $4$                       \\
        Number of pilots for meta-updates while meta-training           & $N_\tau^\text{te}$            & $3000$                & $4$                       \\
        Number of pilots for frame-specific updates while meta-testing  & $N_*^\text{tr}$               & $8$                   & $4$                       \\
        Number of symbols with each channel states while meta-testing   & $N_*^\text{te}$               & $4000$                & $1000$                    \\
        Number of inner SGD updates while meta-training                 & $I$                           & $2$                   & $2$                       \\
        Number of inner SGD updates while meta-testing                  & $I_*$                         & $200$                 & $2$                       \\
        \changemade{Number of meta-updates while meta-training}         & \changemade{$I_\text{meta}$}  & \changemade{$200$}    & \changemade{no. of tasks $t$}          \\
        Ensemble size (for Bayesian framework only)                     & $R$                           & $100$                 & $100$                     \\ 
        Assumed precision of equalizer                                  & $\beta$                       & -                     & $150$                     \\
        Number of frames forming the initial experience                 & $t_\text{init}$               & -                     & $3$                       \\
        Number of meta-training iterations                              & -                             & $200$                 & $100$                     \\
        Number of meta-testing frames averaged over                     & -                             & $50$                  & $100$                    \\ \bottomrule
    \end{tabular}
    \vspace{0.2cm}
\end{table*}

%%%%%%%%%%%%%%%%%%%%%%%%%%%%%%%%%%%%%%%%%%%%%%%%%%%%%%%%%%%%%%%%%%%%%%%%%%%%%%%%%%%%%%
%%%%%%%%%%%%%%%%%%%%%%%%%%%%        Subsection            %%%%%%%%%%%%%%%%%%%%%%%%%%%%
%%%%%%%%%%%%%%%%%%%%%%%%%%%%%%%%%%%%%%%%%%%%%%%%%%%%%%%%%%%%%%%%%%%%%%%%%%%%%%%%%%%%%%
\section{Experiments Details}\label{appendix: Experiments Details}

Table \ref{tab: parameters used in this paper} summarizes the parameters used for the numerical experiments in Sec.~\ref{sec: Experiments} for demodulation and equalization. \changemade{Throughout the simulations, we used $\mathtt{PyTorch}$ \cite{paszke2017automatic} adopting $\mathtt{autograd}$'s option $\mathtt{create\_graph=True}$ to allow the computational graph to calculate second-order derivatives.
}

For the demodulation problem in Sec.~\ref{sec: Experiments Demodulation} (Figs.~\ref{fig: SER_vs_num_mtr_frames}~--~\ref{fig: reliability_diagrams}), the complex input space $\mathcal{Y}=\mathbb{C}$ is treated as a two-dimensional real vector space $\mathbb{R}^2$ when is fed into the neural network demodulator. The KL term in \eqref{eq: F_hat_D_tau(varphi,xi) def} is suppressed by a multiplicative coefficient of 0.1, as a means to emphasize the average log-likelihood term should have over the prior. This is an approach known as generalized Bayesian inference \cite{knoblauch2019generalized, jose2021free}. To handle the discrepancy in the number of pilots for adaptation during meta-training and meta-testing, i.e., $N_*^\text{tr}>N_\tau^\text{tr}$, we consider the following strategy akin to burn-in phase \cite{welling2011bayesian} during meta-testing as done in \cite{park2021fewpilots}: \emph{(i)} start with $I$ updates using learning rate $\eta$ utilizing $N_\tau^\text{tr}$ pilots among the available $N_*^\text{tr}$ pilots; \emph{(ii)} then, additional $I_*- I$ updates are performed with reduced learning rate ($5\%$ of the original learning rate) with all available $N_*^\text{tr}$ pilots. This strategy becomes particularly useful in practical scalable systems in which the number of pilots may change depending on the deployment environments. 

As for the equalization setting in Sec.~\ref{sec: Experiments Equalization} (Figs.~\ref{Fig: scoring function of active}~--~\ref{fig: meta-test loss vs. num meta-training frames}), we observe that reinitializing the hyperparameter $\xi$ to a random value at each data acquisition iteration benefits meta-training in practice. While using the previous iteration's optimized hyperparameter vector $\xi$ as the starting point for the current iteration is useful in reducing the computational complexity \cite{finn2019online, park2021fewpilots}, we found it beneficial not to do so in our equalization problem to avoid meta-overfitting especially in the few-frames (e.g., $10$ frames) regime of interest.

%%%%%%%%%%%%%%%%%%%%%%%%%%%%%%%%%%%%%%%%%%%%%%%%%%%%%%%% end of appendix

\bibliographystyle{IEEEtran} 
\bibliography{my_bib.bib} 
\newpage

\begin{IEEEbiography}[{\includegraphics[width=1in,height=1.25in,clip,keepaspectratio]{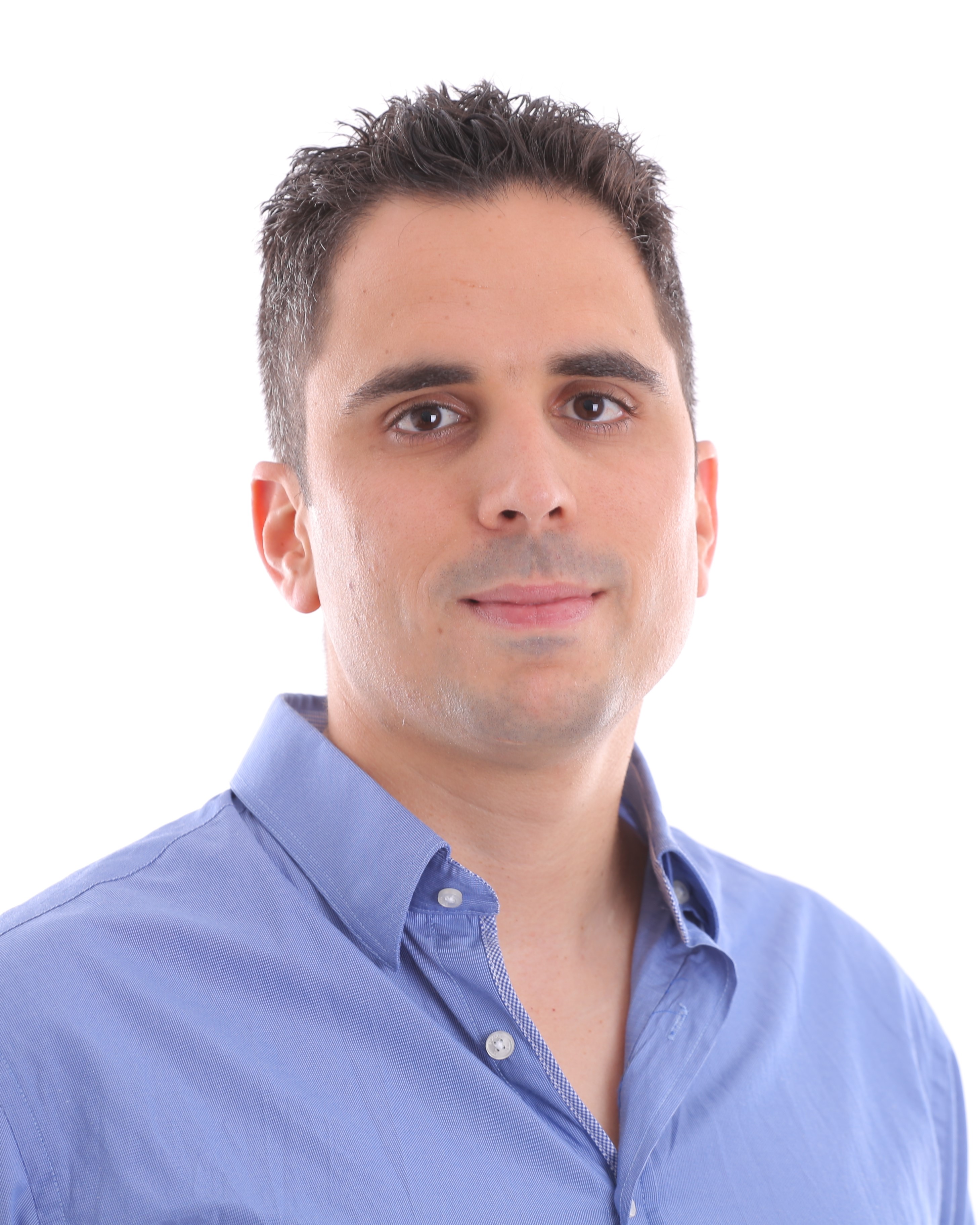}}]{Kfir M. Cohen} (Student Member, IEEE) is pursuing his Ph.D. in the King's Communications, Learning and Information Processing lab at the Department of Engineering of King's College London (KCL). He received B.Sc. (summa cum laude) and M.Sc. degrees in 2006 and 2013, respectively, both in the Electrical Engineering Faculty of the Technion -- Israel Institute for Technology, Haifa, Israel. He served for 15 years in different R\&D roles. Before joining KCL, his last position was as a communication signal processing engineer. His research interests are Bayesian and reliable machine learning, signal processing, as well as their applications to communications systems.
\end{IEEEbiography}

\begin{IEEEbiography}[{\includegraphics[width=1in,height=1.25in,clip,keepaspectratio]{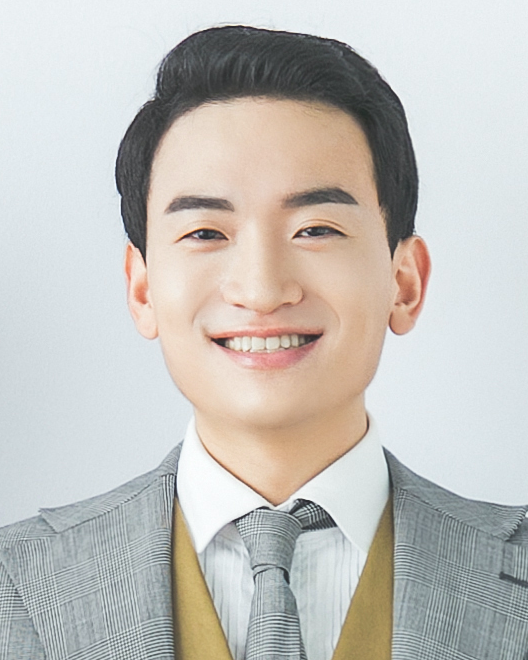}}]{Sangwoo Park}  (Member, IEEE) received his B.S. degree in physics in 2014; M.S.E and Ph.D. degrees in electrical engineering in 2016 and 2020, all from Korea Advanced Institute of Science and Technology (KAIST), Daejeon, Korea. He is currently a research associate in the Department of Engineering, King's Communications, Learning and Information Processing lab, King's College London, United Kingdom. His research interests lie in practical, reliable AI and its application for wireless communication systems and quantum information processing.
\end{IEEEbiography}

\begin{IEEEbiography}[{\includegraphics[width=1in,height=1.25in,clip,keepaspectratio]{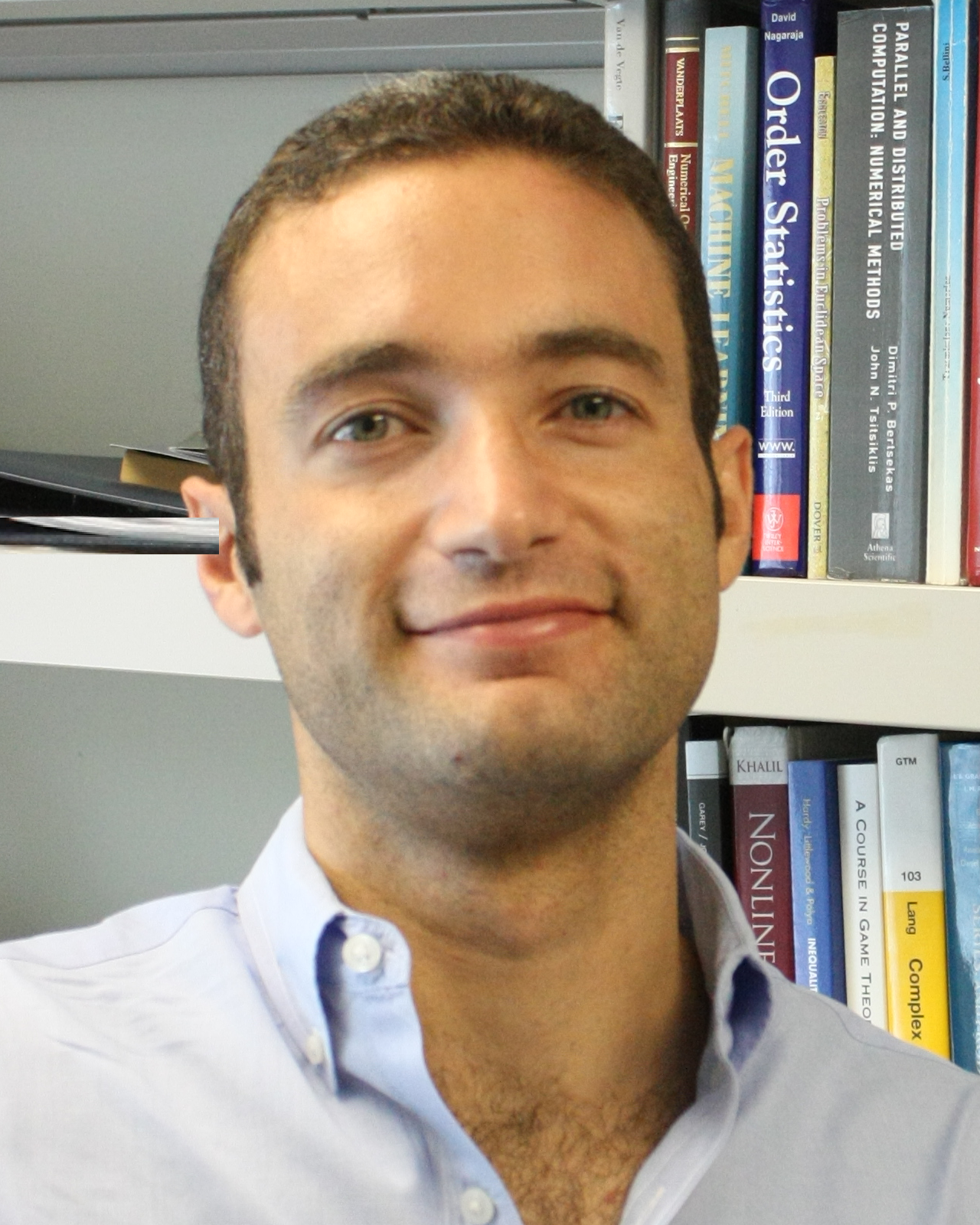}}]{Osvaldo Simeone}  (Fellow, IEEE) is a Professor of Information Engineering with the Centre for Telecommunications Research at the Department of Engineering of King's College London, where he directs the King's Communications, Learning and Information Processing lab. He received an M.Sc. degree (with honors) and a Ph.D. degree in information engineering from Politecnico di Milano, Milan, Italy, in 2001 and 2005, respectively. From 2006 to 2017, he was a faculty member of the Electrical and Computer Engineering (ECE) Department at New Jersey Institute of Technology (NJIT), where he was affiliated with the Center for Wireless Information Processing (CWiP). His research interests include information theory, machine learning, wireless communications, neuromorphic computing, and quantum machine learning. Dr Simeone is a co-recipient of the 2022 IEEE Communications Society Outstanding Paper Award, the 2021 IEEE Vehicular Technology Society Jack Neubauer Memorial Award, the 2019 IEEE Communication Society Best Tutorial Paper Award, the 2018 IEEE Signal Processing Best Paper Award, the 2017 JCN Best Paper Award, the 2015 IEEE Communication Society Best Tutorial Paper Award and of the Best Paper Awards of IEEE SPAWC 2007 and IEEE WRECOM 2007. He was awarded an Open Fellowship by the EPSRC in 2022 and a Consolidator grant by the European Research Council (ERC) in 2016. His research has been also supported by the U.S. National Science Foundation, the European Commission, the European Research Council, the Vienna Science and Technology Fund, the European Space Agency, as well as by a number of industrial collaborations including with Intel Labs and InterDigital. He is the Chair of the Signal Processing for Communications and Networking Technical Committee of the IEEE Signal Processing Society and of the UK \& Ireland Chapter of the IEEE Information Theory Society. He is currently a Distinguished Lecturer of the IEEE Communications Society, and he was a Distinguished Lecturer of the IEEE Information Theory Society in 2017 and 2018. Dr Simeone is the author of the textbook "Machine Learning for Engineers" published by Cambridge University Press, four monographs, two edited books, and more than 180 research journal and magazine papers. He is a Fellow of the IET, EPSRC, and IEEE.
\end{IEEEbiography}

\begin{IEEEbiography}[{\includegraphics[width=1in,height=1.25in,clip,keepaspectratio]{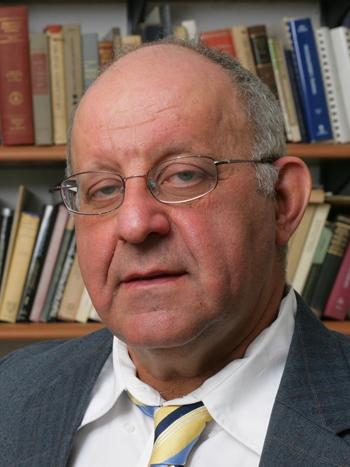}}]{Shlomo Shamai (Shitz)} (Life Fellow, IEEE) is with the Department of Electrical Engineering, Technion—Israel Institute of Technology, where he is a Technion Distinguished Professor, and holds the William Fondiller Chair of telecommunications. Dr. Shamai is an URSI Fellow, a member of the Israeli Academy of Sciences and Humanities, and a Foreign Member of the U.S. National Academy of Engineering. He was a recipient of the 2011 Claude E. Shannon Award, the 2014 Rothschild Prize in Mathematics/Computer Sciences and Engineering, and the 2017 IEEE Richard W. Hamming Medal. He was a co-recipient of the 2018 Third Bell Labs Prize for Shaping the Future of Information and Communications Technology. He was also a recipient of numerous technical and paper awards and recognitions of the IEEE (Donald G. Fink Prize Paper Award), Information Theory, Communications and Signal Processing Societies and EURASIP. He is listed as a Highly Cited Researcher (computer science) for the years 2013/4/5/6/7/8. He has served as an Associate Editor for the Shannon Theory of the IEEE Transactions on Information Theory and has also served twice on the Board of Governors of the Information Theory Society. He has also served on the Executive Editorial Board of the IEEE Transactions on Information Theory, the IEEE Information Theory Society Nominations and Appointments Committee, and the IEEE Information Theory Society (Shannon Award Committee).
\end{IEEEbiography}

\vfill

\end{document}